\documentclass[11pt]{article}
\usepackage[margin=1.1in]{geometry}

\usepackage[utf8]{inputenc}

\usepackage{hyperref}
\usepackage{enumitem}
\usepackage{graphicx}
\graphicspath{{figs/}}
\usepackage{multirow}
\usepackage{booktabs}
\usepackage{tabularx}
\usepackage[title]{appendix}

\usepackage{amsfonts}       
\usepackage{amsmath}       
\usepackage{nicefrac}       
\usepackage{microtype}      
\usepackage{amsthm}
\usepackage{bm}
\usepackage{stmaryrd}
\usepackage{mathtools}
\usepackage{shortcuts}
\usepackage{authblk}

\begin{document}

\title{Robust learning from corrupted EEG with dynamic spatial filtering}

\author[1,2]{Hubert Banville\thanks{correspondence: hubert.jacob-banville@inria.fr}}
\author[2]{Sean U.N. Wood}
\author[2]{Chris Aimone}
\author[1,3]{Denis-Alexander Engemann\thanks{joint senior authors}}
\author[1]{Alexandre Gramfort$^\dagger$}
\affil[1]{Université Paris-Saclay, Inria, CEA, Palaiseau, France}
\affil[2]{InteraXon Inc., Toronto, Canada}
\affil[3]{Max Planck Institute for Human Cognitive and Brain Sciences, Department of Neurology, Leipzig, Germany}

\maketitle
\begin{abstract}
Building machine learning models using EEG recorded outside of the laboratory setting requires methods robust to noisy data and randomly missing channels. This need is particularly great when working with sparse EEG montages (1-6 channels), often encountered in consumer-grade or mobile EEG devices. Neither classical machine learning models nor deep neural networks trained end-to-end on EEG are typically designed or tested for robustness to corruption, and especially to randomly missing channels.
While some studies have proposed strategies for using data with missing channels, these approaches are not practical when sparse montages are used and computing power is limited (\eg wearables, cell phones).
To tackle this problem, we propose dynamic spatial filtering (DSF), a multi-head attention module that can be plugged in before the first layer of a neural network to handle missing EEG channels by learning to focus on good channels and to ignore bad ones.
We tested DSF on public EEG data encompassing $\sim$4,000 recordings with simulated channel corruption and on a private dataset of $\sim$100 at-home recordings of mobile EEG with natural corruption.
Our proposed approach achieves the same performance as baseline models when no noise is applied, but outperforms baselines by as much as 29.4\% accuracy when significant channel corruption is present.
Moreover, DSF outputs are interpretable, making it possible to monitor channel importance in real-time.
This approach has the potential to enable the analysis of EEG in challenging settings where channel corruption hampers the reading of brain signals.
\end{abstract}

\noindent{\textbf{Highlights}\/}
\begin{itemize}
    \setlength{\itemsep}{0pt}
    \setlength{\parskip}{0pt}
    \item We propose a method to handle data corruption in EEG recorded with very few channels.
    \item We design an attention-based neural network architecture to reweight EEG channels on a window-by-window basis according to their relevance given a predictive task.
    \item We validated the method on two large clinical EEG datasets with simulated corruption and on one mobile EEG dataset recorded in at-home settings with naturally occurring channel corruption. 
    \item When significant loss of channels occurs, our method systematically outperforms traditional noise-handling strategies in the context of pathology detection and sleep staging classification tasks.
\end{itemize}

\noindent{\textbf{Keywords}\/}
Electroencephalography, mobile EEG, deep learning, machine learning, noise robustness, sleep staging, pathology detection

\section{Introduction}

Electroencephalography (EEG) enables investigations into brain function and health in an economical manner and for a wide array of purposes, including sleep monitoring, pathology screening, neurofeedback, brain-computer interfacing and anaesthesia monitoring \cite{mihajlovic2014wearable,dhindsa2017filter,kreuzer2017eeg,johnson2020advancing,hohmann2020mynd,krigolson2021using}.
Thanks to recent advances in mobile EEG technology, these applications can now be more easily translated from the lab and clinic to contexts such as at-home or ambulatory assessments.
This carries the potential of democratizing EEG applications and revolutionizing the study of brain health in real-world settings.
However, in these new settings, the number of electrodes available is often limited and signal quality is much harder to control.
Moreover, with the increasing availability of these devices, the amount of data generated now exceeds the capacity of human experts (\eg neurologists, sleep technicians, etc.) to analyze and manually annotate every single recording, as is traditionally done in research and clinical settings.
Novel methods facilitating clinical and research applications in real-world settings, especially with sparse EEG montages, are therefore needed.

The use of machine learning for automating EEG analysis has been the subject of much research in recent decades \cite{lotte2007review,roy2019deep}.
However, state-of-the-art EEG prediction pipelines are generally benchmarked on datasets recorded in well-controlled conditions that are relatively clean when compared to data from mobile EEG.
As a result, it is unclear how models designed for laboratory data will cope with signals encountered in real-world contexts.
This is especially critical for mobile EEG recordings that may contain a varying number of usable channels as well as overall noisier signals, in contrast to most research- and clinical-grade recordings.
In addition, the difference in number of channels between research and mobile settings also means that interpolating bad channels offline (as is commonly done in recordings with dense electrode montages) is likely to fail on mobile EEG devices given their limited spatial information.
It is an additional challenge that the quality of EEG data is not static but can vary significantly within a given recording. This suggests that predictive models should handle noise dynamically.
Ideally, not only should machine learning pipelines produce predictions that are robust to (changing) sources of noise in EEG, but they should also do so in a way that is interpretable.
For instance, if noise is easily identifiable, corrective action can be quickly taken by experimenters or users during a recording.

It is important to consider that not all sources of noise affect EEG recordings in the same way \cite{hari2017meg}.
Physiological artifacts are large electrical signals that are generated by current sources outside the brain such as heart activity, eye or tongue movement, muscle contraction, sweating, etc.
Depending on the EEG electrode montage and the setting of the recording (\eg eyes open or closed), these artifacts can be more or less disruptive to measuring the brain activity of interest.
Movement artifacts, on the other hand, are caused by the relative displacement of EEG electrodes with respect to the scalp, with sharp deflections being created in the affected electrodes during movement.
If an electrode cannot properly reconnect with the skin after the movement (or was not correctly set up initially), its reading will likely contain little or no physiological information and instead pick up instrumentation and environmental noise (\eg from the electronic circuit or powerful electromagnetic sources present around the recording equipment).
We use the terms ``corrupted'' or ``missing'' channels interchangeably to refer to these channels that are strongly contaminated by noise.
Importantly, because these sources of noise are characterized by widely different morphologies and spatial distributions, noise handling techniques typically need to be designed with a specific type of noise in mind.
For instance, channel corruption is likely to occur more often in mobile EEG recordings than in controlled laboratory settings where trained experimenters can monitor and remedy bad electrodes during the recording.

In this paper, we propose and benchmark an attention mechanism module designed to handle corrupted channel data, based on the concept of ``scaling attention''~\cite{hu2018squeeze,woo2018cbam}.
This module can be inserted before the first layer of any convolutional neural network architecture in which activations have a spatial dimension \cite{schirrmeister2017deep,lawhern2018eegnet,chambon2018deep},
and then be trained end-to-end for the prediction task at hand.

The rest of the paper is structured as follows. 
Section~\ref{sec:methods} presents an overview of the EEG noise handling literature, then describes the attention module and denoising procedure proposed in this study.
The neural architectures, baseline methods and data used in our experiments are introduced in Section~\ref{sec:experiments}.
Next, Section~\ref{sec:results} reports the results of our experiments on sleep and pathology EEG datasets.
Lastly, we examine related work in Section~\ref{sec:related_work} and discuss the results in Section~\ref{sec:discussion}.

\section{Methods}
\label{sec:methods}

\subsection{State-of-the-art approaches to noise-robust EEG processing}
\label{subsec:sota_denoising}

Existing strategies for dealing with noisy data can be divided into three categories (Table~\ref{tab:denoising-approaches}): (1)~ignoring or rejecting noisy segments, (2)~implicit denoising, \textit{i.e.}, methods that allow models to work despite noise, and (3)~explicit denoising, \textit{i.e.}, methods that rely on a separate preprocessing step to handle noise or missing channels before inference.
We now discuss existing methods employing these strategies in more detail.

The simplest way to deal with noise in EEG is to assume that it is negligible or to simply discard bad segments \cite{roy2019deep}.
For instance, a manually selected amplitude or variance threshold \cite{manor2015convolutional,hefron2018cross,wang2018data} or a machine learning pipeline trained to recognize artifacts \cite{dhindsa2017filter} can be used to identify noisy segments to be ignored.
This approach, though commonplace, is ill-suited to mobile EEG settings where noise cannot be assumed to be negligible, but also to online applications where model predictions need to be continuously available.
Moreover, this approach is likely to discard windows due to a small fraction of bad electrodes, potentially losing usable information from other channels.

Implicit denoising approaches can be used to design noise-robust processing pipelines that do not contain a specific noise handling step.
A first group of implicit denoising approaches uses representations of EEG data that are robust to missing channels.
For instance, multichannel EEG can be transformed into topographical maps (topomaps) to increase robustness to the absence of a small number of channels.
Typically, a single topomap is extracted per frequency band, and the resulting images are stacked into a single input image.
This representation is then fed into a standard convolutional neural network (ConvNet) architecture, which can learn to focus on the frequencies where the signal-to-noise ratio (SNR) is better.
While this approach can gracefully handle missing channels in dense montages (\eg 16 to 64 channels in \cite{bashivan2015learning,thodoroff2016learning,hagad2019deep}), it is likely to perform poorly on sparse EEG montages (\eg 4 channels) as spatial interpolation might fail if channels are missing.
Moreover, this approach requires computationally demanding preprocessing and feature extraction steps, undesirable in online and low-computational resources contexts.
In the traditional machine learning setting, Sabbagh \textit{et al.} \cite{sabbagh2020predictive} similarly showed that representing input windows as covariance matrices and using Riemannian geometry-aware models did not require common noise correction steps to reach high performance on a brain age prediction task.
However, the robustness of this approach has not been evaluated on sparse montages.
Also, its integration into neural network architectures is not straightforward with geometry-aware deep learning remaining an active field of research \cite{bronstein2018geometric}.
Signal processing techniques can also be used to promote invariance to certain types of noise. 
For instance, the Lomb-Scargle periodogram can be used to extract spectral representations that are robust to missing samples \cite{li2015feature,chu2018decoding}.
However, this approach is not useful if channels are completely missing.
Finally, implicit denoising can be achieved with traditional machine learning models that are inherently robust to noise.
For instance, random forests trained on handcrafted EEG features were shown to be notably more robust to low SNR inputs than univariate models on a state of consciousness prediction task \cite{engemann2018robust}.
Although promising, this approach suffers from the limitations brought about by its feature engineering step, as features (1) rely heavily on domain knowledge, (2) might not be optimal to the task, and (3) require an additional processing step which can be prohibitive in limited resource contexts.

Multiple studies have adopted approaches to explicitly handle noise by automatically correcting corrupted signals or predicting missing or additional channels from the available ones.
Popular spatial projection approaches aim at projecting the input signals to a noise-free subspace before projecting the signals back into channel-space, \eg using independent component analysis (ICA) \cite{jung1998extended,mammone2011automatic,winkler2011automatic} or principal components analysis (PCA) \cite{UusitaloIlmoniemi1997,kothe2016artifact}.
In addition to introducing an additional preprocessing step (and thus additional computational resources), these approaches are again ill-suited to sparse montages settings.
Also, because explicit denoising is decoupled from the learning task, discriminative information is likely to be discarded during preprocessing.
The fact that preprocessing is done independently from the supervised learning task or the statistical testing procedure actually makes the selection of preprocessing parameters (\eg number of good components) challenging.
Motivated by this problem of manual parameter selection, fully automated denoising pipelines have been proposed.
For instance, FASTER \cite{nolan2010faster} and PREP \cite{bigdely2015prep} both combine artifact correction, noise removal and bad channel interpolation into a single automated pipeline.
Autoreject \cite{jas2017autoreject} is another recently developed pipeline that uses cross-validation to automatically select amplitude thresholds to use for rejecting windows or flagging bad channels to be interpolated window-wise.
These approaches are well-suited to offline analyses where the morphology of the signals is of interest, however they are typically computationally demanding and are also decoupled from the statistical modeling.
Additionally, it is unclear how interpolation can be applied when using montages that do not share a single reference, as is often the case in \eg polysomnography \cite{berry2012aasm} and epilepsy monitoring \cite{rosenzweig2014beyond}.

Finally, generic machine learning models have been proposed to recover missing or corrupted channels.
For instance, generative adversarial networks (GANs) have previously been trained to recover dense EEG montages from a few electrodes \cite{corley2018deep,svantesson2020virtual}.
Other similar methods have been proposed, \eg using long short-term memory (LSTM) neural networks \cite{paul2020prediction}, autoencoders \cite{el2019gate}, or tensor decomposition and compressed sensing \cite{ramakrishnan2016reconstruction,sole2018brain}.
However, these methods postulate that the identity of corrupted or missing channels is known ahead of time,
which is a non-trivial assumption in practice.

In contrast to the existing literature on channel corruption handling in EEG, we introduce an interpretable end-to-end denoising approach that can learn implicitly to work with corrupted sparse EEG data, and that does not require additional preprocessing steps.

\begin{table}[]
\caption{Existing methods for dealing with noisy EEG data.}
\small
\renewcommand\arraystretch{1.3}
\label{tab:denoising-approaches}
\begin{tabular}{@{}p{0.15\linewidth}p{0.2\linewidth}p{0.3\linewidth}p{0.35\linewidth}@{}}
\toprule
 &
  Approach &
  Examples &
  Notes \\
\midrule
\textbf{Ignore or} \newline \textbf{reject noise} &
  No denoising &
  \cite{schirrmeister2017deep,lawhern2018eegnet,li2019channel,schirrmeister2017pathology,gemein2020machine,supratak2017deepsleepnet,guillot2020dreem,phan2019seqsleepnet,phan2020xsleepnet} &
  Might not work in real-life applications (out of the lab/clinic) \\
 &
  Removing bad epochs &
  \cite{manor2015convolutional,dhindsa2017filter,hefron2018cross,wang2018data} &
  Doesn’t allow online predictions; Might discard useful information \\
\midrule
\textbf{Implicit} \newline \textbf{denoising} &
  Robust input representations &
  Covariance matrices in Riemannian tangent space \cite{sabbagh2020predictive} &
  Might not work if too few channels available \\
 &
   &
  Topomaps \cite{bashivan2015learning,thodoroff2016learning,hagad2019deep} &
  Expensive preprocessing step; Might not work if too few channels available \\
 &
  Robust signal processing techniques &
  Lomb-Scargle periodogram \cite{li2015feature,chu2018decoding} &
  Only useful for missing samples, not missing channels \\
 &
  Robust machine learning classifiers &
  Handcrafted features and random forest \cite{engemann2018robust} &
  Requires feature engineering step \\
\midrule
\textbf{Explicit} \newline \textbf{denoising} &
  Spatial projection-based approaches &
  Signal Space Separation (SSS) for MEG \cite{taulu2004suppression} &
  Might not work if too few channels available; Additional preprocessing step; Preprocessing might discard important information for learning task \\
 &
   &
  ICA-based denoising \cite{jung1998extended,mammone2011automatic,winkler2011automatic} &
   \\
 &
  Automated correction &
  Autoreject \cite{jas2017autoreject}, FASTER \cite{nolan2010faster}, PREP \cite{bigdely2015prep} &
  Expensive preprocessing step \\
 &
  Model-based interpolation/ reconstruction &
  Deep learning-based superresolution (GAN, LSTM, AE, etc.) \cite{han2018feasibility,kwon2019super,corley2018deep,svantesson2020virtual,el2019gate} &
  Separate training step; Additional inference step to reconstruct at test time; Requires separate procedure to detect corrupted channels \\
 &
   &
  Tensor decomposition, compressed sensing \cite{sole2018brain,ramakrishnan2016reconstruction} & \\
\midrule
\textbf{Interpretable denoising} &
   \textbf{Channel corruption-invariant architecture} &
   \textbf{Dynamic Spatial Filtering (this work)} &
   \textbf{Trained end-to-end, no additional preprocessing, interpretable, works with sparse montages} \\
\bottomrule
\end{tabular}
\end{table}

\subsection{Dynamic spatial filtering: Second-order attention for learning on noisy EEG signals}
\label{subsec:dsf_description}

The key goal behind dynamic spatial filtering (DSF) is to help neural networks focus on the most important channels, at each time instant, given a specific machine learning task on EEG.
To do so, we introduce a spatial attention mechanism that dynamically reweights channels according to their predictive power.
This idea is inspired by recent developments in attention mechanisms, most specifically the ``scaling attention'' approach proposed in computer vision \cite{hu2018squeeze,woo2018cbam}.
Notably, DSF leverages second-order information, \textit{i.e.}, spatial covariance, to capture dependencies between EEG channels.
In this section, we detail the learning problem under study, the proposed attention architecture and a data augmentation transform designed to help train noise-robust models.

\paragraph*{Notation}
We denote by $\intset{q}$ the set $\{1, \ldots, q\}$.
The index $t$ refers to time indices in the multivariate time series $S \in \bbR^{C \times M}$, where $M$ is the number of time samples and $C$ is the number of EEG channels.
$S$ is further divided into non-overlapping windows $X \in \bbR^{C \times T}$ where $T$ is the number of time samples in the window.
We denote by $y \in \mathcal{Y}$ the target used in the learning task.
Typically, $\mathcal{Y}$ is $\intset{L}$ for a classification problem with $L$ classes.

We perform experiments in the supervised classification setting.
A model $f_{\Theta} : \mathcal{X} \rightarrow \mathcal{Y}$ with parameters $\Theta$ is trained to predict the class $y$ of EEG windows $X$.
For instance, $f_\Theta$ can be implemented as a convolutional neural network.
For this, we train $f_\Theta$ to minimize the loss $\mathcal{L}$, \eg the categorical cross-entropy loss, over the example-label pairs $(X_i,y_i)$:

\begin{equation}
    \hat{f_\Theta} = \arg\min_{\Theta} \mathbb{E}_{X_i, y_i \in \mathcal{X} \times \mathcal{Y}} [\mathcal{L}(f_\Theta(X_i),y_i)] \enspace .
\end{equation}

In particular, we are interested in the performance of $f_\Theta$ when random channels are corrupted and more specifically when channel corruption occurs at test time (\textit{i.e.}, when training data is mostly clean).
Toward this goal, we insert an attention-based module $m_{\mathrm{DSF}}: \bbR ^ {C \times T} \rightarrow \bbR ^ {C' \times T}$ into $f_\Theta$ which performs a (fixed) transformation $\Phi(X)$ to extract relevant spatial information from $X$, followed by a reweighting mechanism for the input signals.

\begin{figure}
    \centering
    \includegraphics[width=\textwidth]{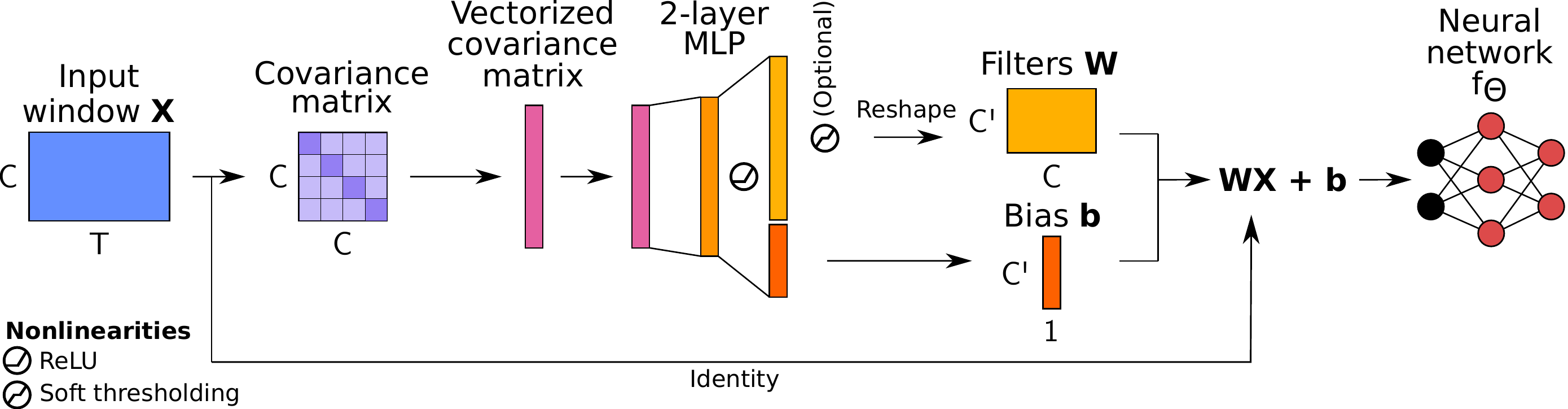}
    \caption{Visual description of the Dynamic Spatial Filtering (DSF) attention module. An input window $\bX$ with $C$ spatial channels is processed by a 2-layer MLP to produce a set of $C'$ spatial filters $\bW$ and biases $\bb$ that dynamically transform the input $\bX$. This allows the subsequent layers of a neural network to ignore bad channels and focus on the most informative ones.}
    \label{fig:dsf_arch}
\end{figure}

In order to implicitly handle noise in neural network architectures, we design an attention module where second-order information is extracted from the input and used to predict weights of a linear transformation of the input EEG channels, that are optimized for the learning task (Fig.~\ref{fig:dsf_arch}).
Applying such linear transforms to multivariate EEG signals is commonly referred to as ``spatial filtering'', a technique that has been widely used in the field of EEG \cite{makeig1996independent,mcfarland1997spatial,parra2005recipes,blankertz2007optimizing,de2008denoising,lotte2010regularizing,nikulin2011novel}.
This enables the model to learn to ignore noisy outputs and/or to reweight them, while still leveraging any remaining spatial information. 
We now show how this module can be applied to the raw input $X$.

We define the dynamic spatial filter (DSF) module $m_{\mathrm{DSF}}$ as:

\begin{equation}
    m_{\mathrm{DSF}}(X) = W_{\mathrm{DSF}}(X) X + b_{\mathrm{DSF}}(X)\label{eq:dsf} \enspace ,
\end{equation}
where $W_{\mathrm{DSF}} \in \bbR ^ {C' \times C}$ and $b_{\mathrm{DSF}} \in \bbR ^ {C'}$ are obtained by reshaping the output of a neural network, \eg a multilayer perceptron (MLP), $h_{\Theta_{DSF}}(\Phi(X)) \in \bbR ^ {C' \times (C+1)}$ (see Fig.~\ref{fig:dsf_arch}).
Under this formulation, each row in $W_{\mathrm{DSF}}$ corresponds to a spatial filter that linearly transforms the input signals into another virtual channel.
Here, $C'$ can be set to the number of input spatial channels $C$ or considered a hyperparameter of the attention module\footnote{In which case it can be used to increase the diversity of input channels in models trained on sparse montages ($C' > C$) or perform dimensionality reduction to reduce computational complexity ($C' < C$).}.
When $C' = C$, if the diagonal of $W_{\mathrm{DSF}}$ is 0, $W_{\mathrm{DSF}}$ corresponds to a linear interpolation of each channel based on the $C-1$ others, as is commonly done in the classical EEG literature \cite{perrin1989spherical} (see Appendix~\ref{sec:interp_to_dsf} for an in-depth discussion).
Heavily corrupted channels can be ignored by giving them a weight of 0 in $W_{\mathrm{DSF}}$.
To facilitate this behavior, we can further apply a soft-thresholding element-wise nonlinearity to $W_{\mathrm{DSF}}$:
\begin{equation}
    W'_{\mathrm{DSF}} = \operatorname{sign}(W_{\mathrm{DSF}}) \operatorname{max}(\left | W_{\mathrm{DSF}} \right | - \tau, 0) \enspace ,
\end{equation}
where $\tau$ is a threshold empirically set to 0.1, $\left| \cdot \right |$ is the element-wise absolute value and both the $\operatorname{sign}$ and $\operatorname{max}$ operators are applied element-wise.

In our experiments, the spatial information extracted by the transforms $\Phi(X)$ was either (1) the log-variance of each input channel or (2) the flattened upper triangular part of the matrix logarithm of the covariance matrix of $X$ (see Appendix~\ref{sec:spatial_representation})\footnote{In practice, if a channel is ``flat-lining'' (has only 0s) inside a window and therefore has a variance of 0, its log-variance is replaced by 0. Similarly, if a covariance matrix eigenvalue is 0 when computing the matrix logarithm (see Equation~\ref{eq:logm}), its logarithm is replaced by 0.}.
When reporting results, we denote models as \textit{DSFd} and \textit{DSFm} when DSF takes the log-variance or the matrix logarithm of the covariance matrix as input, respectively.
We further add the suffix ``-st'' to  indicate the use of the soft-thresholding nonlinearity, \eg \textit{DSFm-st}.

Interestingly, the DSF module can be seen as a multi-head attention mechanism \cite{vaswani2017attention} with real-valued attention weights and where each head is tasked with producing a linear combination of the input spatial signals.

Finally, we can inspect the attention given by $m_{\mathrm{DSF}}$ to each input channel by computing the channel contribution metric $\boldsymbol \phi \in \bbR^{C}$ where $\phi_j = \sqrt{\sum_{i=1}^{C'} {W_{ij}}^2}$.
Intuitively, $\boldsymbol\phi$ measures how much each input channel is used by $m_{\mathrm{DSF}}$ to produce the output virtual channels.
This straightforward way of inspecting the functioning of the DSF module facilitates the identification of important or noisy channels.

To further help our models learn to be robust to noise, we design a data augmentation procedure that randomly corrupts channels.
Specifically, channel corruption is simulated by performing a masked channel-wise convex combination of input channels and Gaussian white noise $Z \in \bbR^{C \times T}$:

\begin{equation}
    \label{eq:data_augm}
    \tilde{X} = (1 - \eta) \operatorname{diag}(\boldsymbol{\nu}) X + \eta \operatorname{diag}(\boldsymbol{\nu}) Z + \operatorname{diag}(1 - \boldsymbol{\nu}) X \enspace ,
\end{equation}
where $Z_{i,j} \sim \mathcal{N}(0, \sigma_n ^ 2)$ for $i \in \intset{T}$ and $j \in \intset{C}$, $\eta \in [0, 1]$ controls the relative strength of the noise, and $\boldsymbol{\nu} \in \{0,1\}^C$ is a masking vector that controls which channels are corrupted.
The operator $\operatorname{diag}(x)$ creates a square matrix filled with zeros whose diagonal is the vector $x$.
Here, $\boldsymbol{\nu}$ is sampled from a multinouilli distribution with parameter $p$.
Each window $X$ is individually corrupted using random parameters $\sigma_n \sim \mathcal{U}(20,50)~\mu$V, $\eta \sim \mathcal{U}(0.5,1)$, and a fixed $p$ of 0.5.

\subsection{Computational considerations}

We set the following hyperparameters when training deep neural networks: optimizer, learning rate schedule, batch size, regularization strength (number of training epochs, weight decay, dropout) and parameter initialization scheme.
In all experiments, we used the AdamW optimizer \cite{loshchilov2017decoupled} with $\beta_1=0.9$, $\beta_2=0.999$, a learning rate of $10^{-3}$ and cosine annealing.
The parameters of all neural networks were randomly initialized using uniform He initialization \cite{he2015delving}.
Dropout \cite{srivastava2014dropout} was applied to $f_\Theta$'s fully connected layer at a rate of 50\% and weight decay was applied to the trainable parameters of all layers of both $f_\Theta$ and $h_{\Theta_{DSF}}$.
Moreover, during training, the loss was weighted to optimize balanced accuracy.
Some hyperparameters were tuned on a dataset-specific basis and are described along with the datasets (\textit{i.e.}, weight decay and batch size).

Deep learning and baseline models were trained using a combination of the braindecode \cite{schirrmeister2017deep}, MNE-Python \cite{gramfort2014mne}, PyTorch \cite{paszke2019pytorch},  pyRiemann \cite{barachant2013classification}, mne-features \cite{schiratti2018ensemble} and scikit-learn \cite{pedregosa2011scikit} packages.\footnote{Code used for data analysis can be found at \url{https://github.com/hubertjb/dynamic-spatial-filtering}.}
Finally, deep learning models were trained on 1 or 2 Nvidia Tesla V100 or P4 GPUs for anywhere from a few minutes to 7 hours, depending on the amount of data, early stopping and GPU configuration.

\section{Experiments}
\label{sec:experiments}

\subsection{Downstream tasks}

We studied noise robustness through two common EEG classification downstream tasks: pathology detection and sleep staging.
First, sleep staging, a critical step in sleep monitoring, allows the diagnosis and study of sleep disorders such as apnea and narcolepsy \cite{bathgate2019diagnostic}.
This 5-class classification problem consists of predicting which sleep stage (W (wake), N1, N2, N3 (different levels of sleep) or R (rapid eye movement periods)) an individual is in, in non-overlapping 30-s windows of overnight recordings.
While a large number of machine learning approaches have been proposed to perform sleep staging \cite{motamedi2014signal,chambon2018deep,roy2019deep,phan2020xsleepnet}, the handling of corrupted channels in overnight recordings has not been addressed in a comprehensive manner yet, as channel corruption is less likely to occur in clinical and laboratory settings than in the real-world settings we consider here\footnote{A recent study reported training a neural network on artificially-corrupted sleep EEG data, with a goal similar to ours \cite{jonsson20200447}; however, this study only appears as a Supplement with little information on the methods and results.}.

Second, the pathology detection task aims at detecting neurological conditions such as epilepsy and dementia from an individual's EEG \cite{smith2005eeg,micanovic2014diagnostic}.
In a simplified formulation this gives rise to a binary classification problem where recordings have to be classified as either pathological or non-pathological.
Such recordings are typically carried out in well-controlled settings (\eg in a hospital \cite{obeid2016temple}) where sources of noise can be monitored and mitigated in real-time by experts.
To test pathology detection performance in the context of mobile EEG acquisition, we used a limited set of electrodes, in contrast to previous work \cite{lopez2015automated,schirrmeister2017pathology,gemein2020machine}.

These two tasks are further described in Section~\ref{subsec:data} when discussing the data used in our experiments.

\subsection{Compared methods}
\label{subsec:compared_methods}

We compared the performance of the proposed DSF and data augmentation method to other established approaches.
In total, we contrasted combinations of three machine learning pipelines and three different noise-handling strategies.

We consider the following machine learning pipelines: (1) end-to-end deep learning (with and without the DSF module) from raw signals, (2) filter-bank covariance matrices with Riemannian tangent space projection and logistic regression \cite{barachant2013classification,congedo2017riemannian,lotte2018review,sabbagh2020predictive} (which we refer to as ``Riemann''), and (3) handcrafted features and random forest (RF) \cite{gemein2020machine}.

We used ConvNet architectures as $f_\Theta$ in deep learning pipelines (Appendix~\ref{sec:architectures}).
For pathology detection, we used the ShallowNet architecture from \cite{schirrmeister2017deep} which parametrizes the frequency-band common spatial patterns (FBCSP) pipeline \cite{gemein2020machine}.
We used it without modifying the architecture, yielding a total of 13,482 trainable parameters when $C=6$.
For sleep staging, we used a 3-layer ConvNet which takes 30-s windows as input \cite{chambon2018deep,banville2020uncovering}, with a total of 18,457 trainable parameters when $C=4$ and an input sampling frequency of 100~Hz.
Finally, when evaluating DSF, we added modules $m_{\mathrm{DSF}}$ before the input layer of each neural network.
The input dimensionality of $m_{\mathrm{DSF}}$ depends on the chosen spatial information extraction transform $\Phi(X)$: either $C$ (log-variance) or $C(C+1)/2$ (vectorized covariance matrix).
We fixed the hidden layer size of $m_{\mathrm{DSF}}$ to $C^2$ units, while the output layer size depended on the chosen $C'$.
The DSF modules added between 420 and 2,864 trainable parameters to those of $f_\Theta$ depending on the configuration.

The Riemann pipeline first applied a filter bank to the input EEG, yielding narrow-band signals in the 7 bands bounded by (0.1, 1.5, 4, 8, 15, 26, 35, 49) Hz.
Next, covariance matrices were estimated per window and frequency band using the OAS algorithm \cite{chen2010shrinkage}.
The covariance matrices were then projected into their Riemannian tangent space exploiting the Wasserstein distance to estimate the mean covariance used as the reference point \cite{sabbagh2019manifold,bhatia2018bures}.
The vectorized covariance matrices with dimensionality of $C(C + 1)/2$ were finally z-score normalized using the mean and standard deviation of the training set, and fed to a linear logistic regression classifier.

The handcrafted features baseline, inspired by \cite{gemein2020machine} and \cite{engemann2018robust}, relied on 21 different feature types: mean, standard deviation, root mean square, kurtosis, skewness, quantiles (10, 25, 75 and 90th), peak-to-peak amplitude, frequency log-power bands between (0, 2, 4, 8, 13, 18, 24, 30, 49) Hz as well as all their possible ratios, spectral entropy, approximate entropy, SVD entropy, Hurst exponent, Hjorth complexity, Hjorth mobility, line length, wavelet coefficient energy, Higuchi fractal dimension, number of zero crossings, SVD Fisher information and phase locking value.
This resulted in 63 univariate features per EEG channel, along with ${C \choose 2}$ bivariate features, which were concatenated into a single vector of size $63 \times C + {C \choose 2}$ (\eg 393 for $C=6$).
In the event of non-finite values in the feature representation of a window, we imputed missing values feature-wise using the mean of the feature computed over the training set.
Finally, feature vectors were fed to a random forest model.

When applying traditional pipelines to pathology detection experiments, we aggregated the input representations recording-wise as each recording has a single label (\textit{i.e.}, pathological or not).
To do so, we used the geometric mean on covariance matrices and the median on handcrafted features.
Deep learning models, on the other hand, were trained on non-aggregated windows, but their performance was evaluated recording-wise by averaging the predictions over windows within each recording.
Hyperparameter selection for logistic regression and random forest models is described in Appendix~\ref{sec:hp_opt}.

We combined the machine learning approaches described above with the following noise-handling strategies: (1) no denoising, \textit{i.e.}, models are trained directly on the data without explicit or implicit denoising, (2) Autoreject \cite{jas2017autoreject}, an automated correction pipeline, and (3) data augmentation, which randomly corrupts channels during training.

Autoreject is a denoising pipeline that explicitly handles noisy epochs and channels in a fully automated manner \cite{jas2017autoreject}.
First, using a cross-validation procedure, it finds optimal channel-wise peak-to-peak amplitude thresholds to be used to identify bad channels in each window separately.
If more than $\kappa$ channels are bad, the epoch is rejected.
Otherwise, up to $\rho$ bad channels are reconstructed using the good channels with spherical spline interpolation.
In pathology detection experiments, we allowed Autoreject to reject bad epochs, as classification was performed recording-wise.
For sleep staging experiments however, we did not reject epochs as one prediction per epoch was needed, but still used Autoreject to automatically identify and interpolate bad channels.
In both cases, we used default values for all parameters as provided in the Python implementation\footnote{\url{https://github.com/autoreject/autoreject}}, except for the number of cross-validation folds, which we set to 5.

Finally, data augmentation consists of artificially corrupting channels during training to promote invariance to missing channels.
When training neural networks, the data augmentation transform was applied on-the-fly to each batch.
For feature-based methods, we instead precomputed augmented datasets by applying the augmentation multiple times to each window (10 for pathology detection, 5 for sleep staging), and then extracting features from the augmented windows.

\subsection{Data}
\label{subsec:data}

Approaches were compared on three datasets (Table~\ref{tab:datasets}): for pathology detection on the TUH Abnormal EEG dataset \cite{obeid2016temple} and for sleep staging on both the Physionet Challenge 2018 dataset \cite{ghassemi2018you,goldberger2000physiobank} and an internal dataset of mobile overnight EEG recordings.

\begin{table}[]
\centering
\caption{Description of the datasets used in this study.}
\label{tab:datasets}
\begin{tabular}{lccc}
\toprule
                        & TUAB \cite{lopez2017automated,obeid2016temple} & PC18 (train) \cite{ghassemi2018you, goldberger2000physiobank} & MSD            \\
\midrule
Recording settings      & Hospital         & Sleep clinic        & At-home        \\
\# recordings           & 2,993            & 994                 & 98             \\
\# unique subjects      & 2,329            & 994                 & 67             \\
Sampling frequency (Hz) & 250, 256 or 512  & 200                 & 256            \\
\# EEG channels         & 27 to 36         & 6                   & 4              \\
Reference               & Common average   & M1 or M2            & Fpz            \\
Labels                  & Normal, abnormal & W, N1, N2, N3, R    & W, N1, N2, N3, R \\
\bottomrule
\end{tabular}
\end{table}

\paragraph{TUH Abnormal EEG dataset (TUAB)}

The TUH Abnormal EEG dataset v2.0.0 (TUAB) \cite{lopez2017automated,obeid2016temple} contains 2,993 recordings of 15 minutes or more from 2,329 different patients who underwent a clinical EEG exam in a hospital setting.
Each recording was labeled as ``normal'' (1,385 recordings) or ``abnormal'' (998 recordings) based on detailed physician reports.
Most recordings were sampled at 250~Hz and comprised between 27 and 36 electrodes.
The corpus is already divided into a training and an evaluation set with 2,130 and 253 recordings each.
The mean age across all recordings is 49.3 years (min: 1, max: 96) and 53.5\% of recordings are of female patients.
The TUAB data was preprocessed in the following manner.
The first minute of each recording was cropped to remove noisy data that occurs at the beginning of recordings \cite{gemein2020machine}.
Longer files were cropped such that a maximum of 20 minutes was used from each recording.
Then, 21 channels common to all recordings were selected (Fp1, Fp2, F7, F8, F3, Fz, F4, A1, T3, C3, Cz, C4, T4, A2, T5, P3, Pz, P4, T6, O1 and O2).
EEG channels were downsampled to 100~Hz and clipped at $\pm 800\,\mu V$.
Finally, non-overlapping windows of 6 seconds were extracted, yielding windows of size ($600 \times 21$).
Deep learning models were trained on TUAB with a batch size of 256 and weight decay of 0.01.

\paragraph{Physionet Challenge 2018 dataset (PC18)}
The Physionet Challenge 2018 (PC18) dataset \cite{ghassemi2018you, goldberger2000physiobank} contains recordings from a total of 1,983 different individuals with (suspected) sleep apnea whose EEG, EOG, chin EMG, respiration airflow and oxygen saturation were monitored overnight.
Bipolar EEG channels F3-M2, F4-M1, C3-M2, C4-M1, O1-M2 and O2-M1 were recorded at 200~Hz.
Sleep stage annotations were obtained from 7 trained scorers following the AASM manual \cite{berry2012aasm} (W, N1, N2, N3 and R).
We focused our analysis on a subset of 994 recordings for which these annotations are publicly available.
In this subset of the data, mean age is 55 years (min: 18, max: 93) and 33\% of participants are female.
For PC18, the EEG was first filtered using a 30~Hz FIR lowpass filter with a Hamming window to reject higher frequencies that are not critical for sleep staging \cite{chambon2018deep,aboalayon2016sleep}.
The EEG channels were then downsampled by a factor of two to 100\,Hz to reduce the dimensionality of the input data.
Finally, non-overlapping windows 30 seconds in size ($3000 \times 6$) were extracted.
Experiments on PC18 used a batch size of 64 and weight decay of 0.001.

\paragraph{Muse Sleep Dataset (MSD)}
We lastly tested our approach on mobile EEG data, in which channel corruption is likely to occur naturally.
We used an internal dataset of overnight sleep recordings collected with the Muse S EEG headband from InteraXon Inc. (Toronto, Canada).
This data was collected in accordance with the privacy policy (July 2020) users must agree to when using the Muse headband\footnote{https://choosemuse.com/legal/privacy/} and which ensures their informed consent concerning the use of EEG data for scientific research purposes.
The Muse S is a four-channel dry EEG device (TP9, Fp1, Fp2, TP10, referenced to Fpz), sampled at 256 Hz.
The Muse headband has been previously used for event-related potentials research \cite{krigolson2017choosing}, brain performance assessment \cite{krigolson2021using}, research into brain development \cite{hashemi2016characterizing}, sleep staging \cite{koushik2018real}, and stroke diagnosis \cite{wilkinson2020predicting}, among others.
A total of 98 partial and complete overnight recordings (mean duration: 6.3 h) from 67 unique users were selected from InteraXon's anonymized database of Muse customers, and annotated by a trained scorer following the AASM manual.
Mean age across all recordings is 37.9 years (min: 21, max: 74) and 45.9\% of recordings are of female users.
Preprocessing of MSD data was the same as for PC18, with the following differences: (1) channels were downsampled to 128~Hz, (2) missing values (occurring when the wireless connection is weak and Bluetooth packets are lost) were replaced by linear interpolation using surrounding valid samples, (3) after filtering and downsampling, samples which overlapped with the original missing values were replaced by zeros, and (4) channels were zero-meaned window-wise.
We used a batch size of 64 and weight decay of 0.01 for MSD experiments.


We split the available recordings from TUAB, PC18 and MSD into training, validation and testing, such that recordings used for testing were not used for training or validation. 
For TUAB, we used the provided evaluation set as the test set.
The recordings in the development set were split 80-20\% into a training and a validation set.
Therefore, we used 2,171, 543 and 276 recordings in the training, validation and testing sets.
For PC18, we used a 60-20-20\% random split, meaning there were 595, 199 and 199 recordings in the training, validation and testing sets respectively.
Finally, for MSD, we retained the 17 most corrupted recordings for the test set (Appendix~\ref{sec:msd_corruption_analysis}) and randomly split the remaining 81 recordings into training and validation sets (65 and 16 recordings, respectively).
This was done to emulate a situation where training data is mostly clean, and strong channel corruption occurs unexpectedly at test time.
We performed hyperparameter selection on each of the three datasets using a cross-validation strategy on the combined training and validation sets.  

We repeated training on different training-validation splits (two for PC18, three for TUAB and MSD).
Neural networks and random forests were trained three times per split on TUAB and MSD (two times on PC18) with different parameter initializations.
Training ran for at most 40 epochs or until the validation loss stopped decreasing for a period of a least 7 epochs on TUAB and PC18 (a maximum of 150 epochs with a patience of 30 for MSD, given the smaller size of the dataset).

Finally, balanced accuracy (bal acc), defined as the average per-class recall, was used to evaluate model performance on the sleep staging downstream task, while accuracy was used for pathology detection.
Balanced accuracy was used on sleep data due to important class imbalance.
Specifically, the N2 class is typically much more frequent than other classes. 

\subsection{Evaluation under conditions of noise}

The impact of noise on downstream performance and on the predicted DSF filters was evaluated in three steps.
First, we artificially corrupted the input EEG windows of TUAB and PC18 by using a similar process to our data augmentation strategy (Equation~\ref{eq:data_augm}).
We used the same values for $\eta$, $\sigma$ and $p$, but used a single mask $\boldsymbol{\nu}$ per recording, such that the set of corrupted channels remained the same across a recording.
Before corrupting, we subsampled a few EEG channels to recreate the sparse montage settings of TUAB (Fp1, Fp2, T3, T4, Fz, Cz) and PC18 (F3-M2, F4-M1, O1-M2, O2-M1).
We then analyzed downstream performance under varying noise level conditions.
Second, we ran experiments on real corrupted data (MSD) by training our models on the cleanest recordings and evaluating their performance on the noisiest recordings.
Finally, we analyzed the distribution of DSF filter weights predicted by a subset of the trained models.

\section{Results}
\label{sec:results}

\subsection{Performance of existing methods degrades under channel corruption}
\label{subsec:n_channels_results}

\begin{figure}
    \centering
    \includegraphics[width=\textwidth]{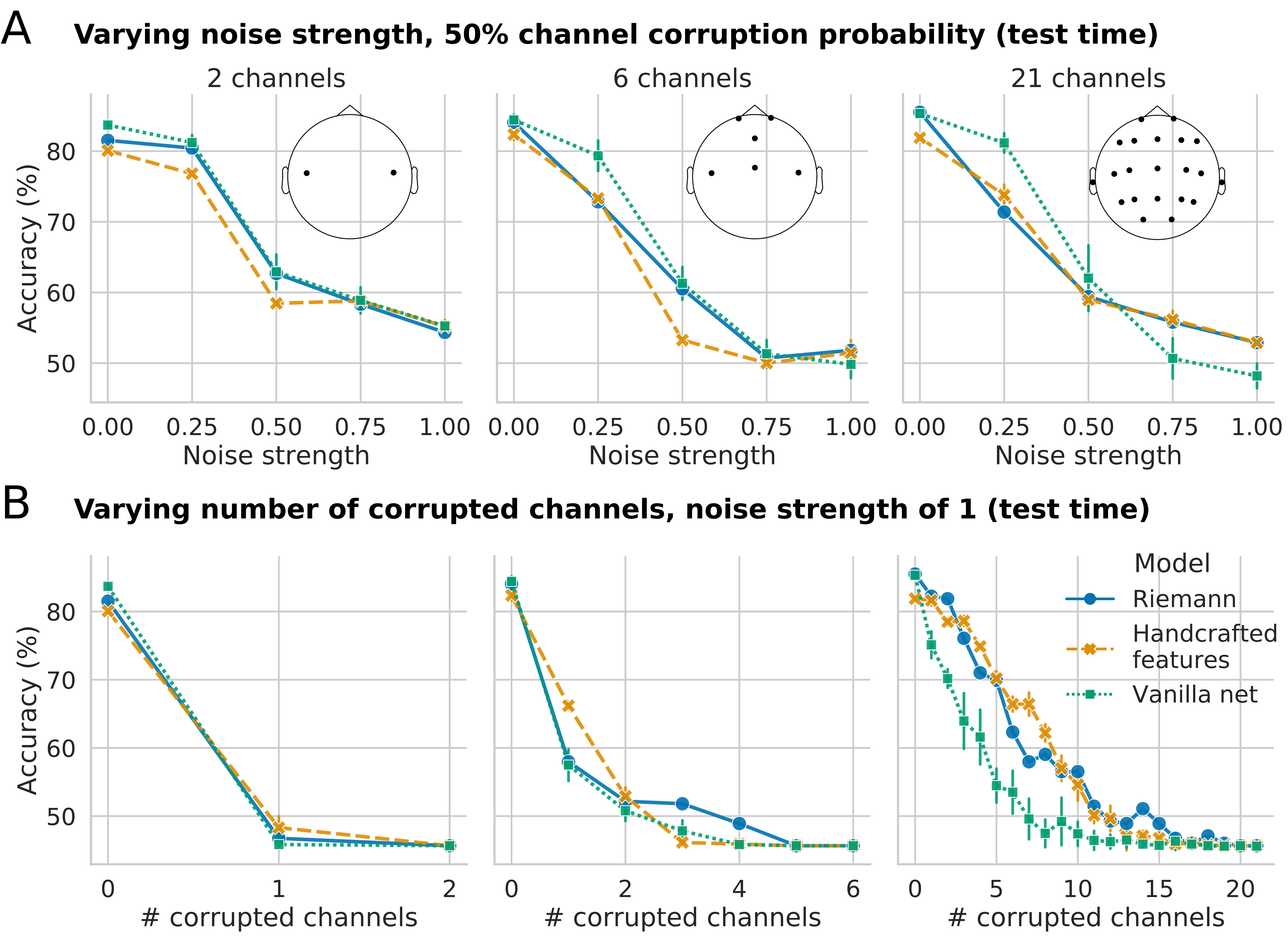}
    \caption{Impact of channel corruption on pathology detection performance of standard models.
    We trained a filter-bank Riemannian geometry pipeline (blue), a random forest on handcrafted features (orange) and a standard ShallowNet architecture (green) on the TUAB dataset, given montages of 2 (T3, T4), 6 (Fp1, Fp2, T3, T4, Fz, Cz) or 21 (all available) channels.
    Performance was then evaluated on artificially corrupted test data under two scenarios: (A) the $\eta$ noise strength parameter was varied given a constant channel corruption probability of 50\%, and (B) the number of corrupted channels was varied given a constant noise strength of 1.
    Error bars show the standard deviation over 3 models for handcrafted features and 6 models for neural networks.
    While traditional feature-based models fared slightly better than a vanilla neural network in some cases (bottom right), adding noise predictably degraded the performance of all three models.
    }
    \label{fig:impact_noise_on_n_channels}
\end{figure}

How do standard EEG classification methods fare against channel corruption?
If channels have a high probability of being corrupted at test time, can noise be compensated for by adding more channels?
To answer these questions, we measured the performance of three baseline approaches (Riemannian geometry, handcrafted features and a ``vanilla'' net, \textit{i.e.}, ShallowNet without attention) trained on a pathology detection task on three different montages as channels were artificially corrupted.
Results are presented in Fig.~\ref{fig:impact_noise_on_n_channels}.

All three baseline methods performed similarly and suffered considerable performance degradation as stronger noise was added (Fig.~\ref{fig:impact_noise_on_n_channels}A) and as more channels were corrupted (Fig.~\ref{fig:impact_noise_on_n_channels}B).
First, under progressively noisier conditions, compensating for corruption by adding more channels did not generally improve performance.
Strikingly, adding channels even hampered the ability of the models to handle noise.
Indeed, the impact of noise was much less significant for 2-channel models than for 6- or 21-channel models.
The vanilla net performed slightly better than the other methods in low noise conditions, however it was less robust to heavy noise when using 21 channels.

Second, when an increasing number of channels was corrupted (B), using denser montages did improve performance, although by a much smaller factor than what might be expected.
For instance, losing one or two channels with the 21-channel models only yielded a minor decrease in performance, while models trained on sparser montages lost as much as 30\% accuracy.
However, even when as many as 15 channels were still available (\textit{i.e.}, six corrupted channels), models trained on 21 channels performed worse than 2- or 6-channel models without any channel corruption, despite having access to much more spatial information on average.
Interestingly, when models were trained on 21 channels, traditional feature-based methods were more robust to corruption than a vanilla net up to a certain point, however this did not hold for sparser montages.

These results suggest that standard approaches cannot handle significant channel corruption at a satisfactory level, even when denser montages are available.
Therefore, better tools are necessary to train noise-robust models.

\subsection{Attention and data augmentation mitigates performance loss under channel corruption}
\label{subsec:denoising_results}

\begin{figure}
    \centering
    \includegraphics[width=\textwidth]{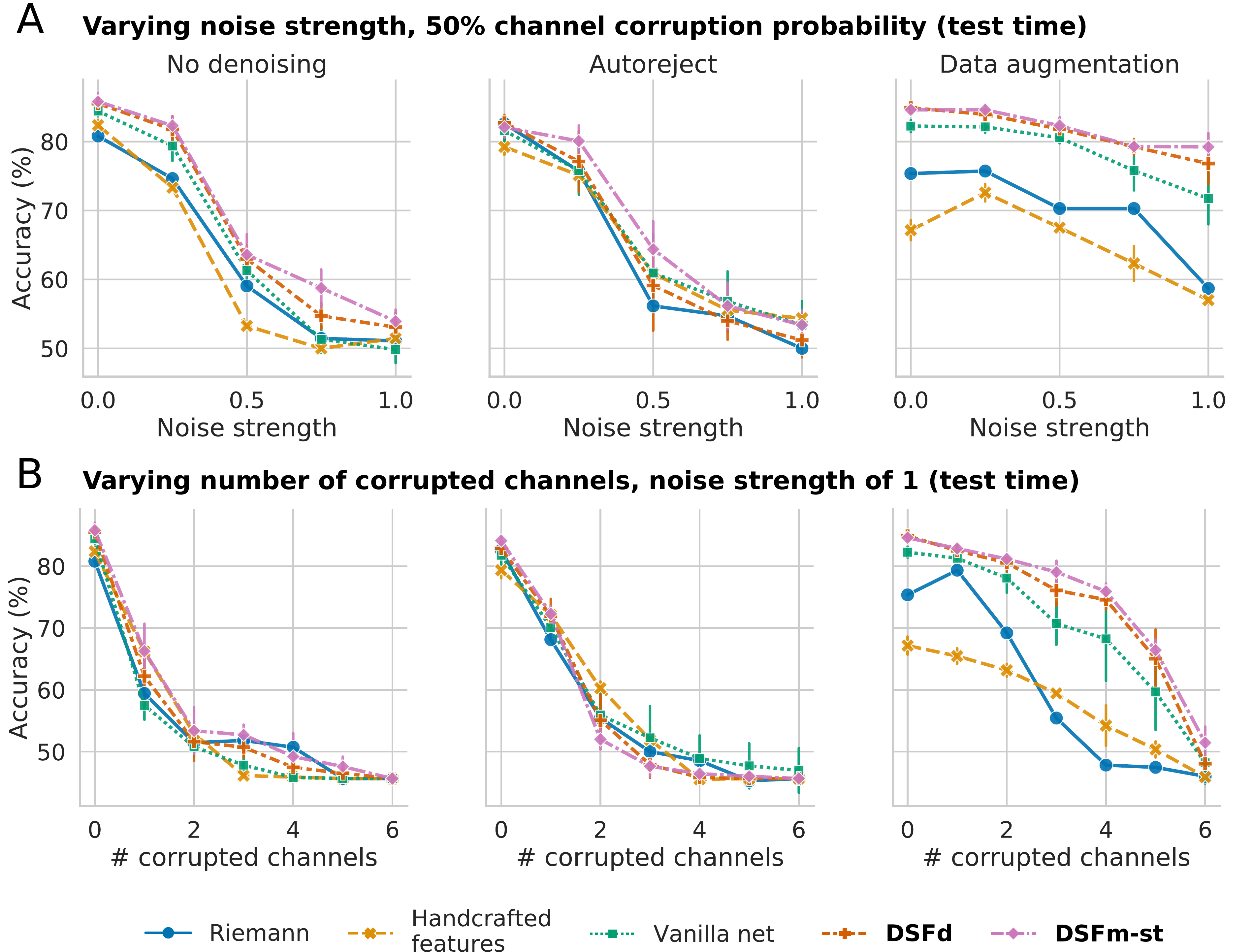}
    \caption{Impact of channel corruption on pathology detection performance for models coupled with (1) no denoising strategy, (2) Autoreject and (3) data augmentation. 
    We compared the per recording accuracy on the TUAB evaluation set (6-channel montage) as (A) the $\eta$ noise strength parameter was varied given a constant channel corruption probability of 50\%, and (B) the number of corrupted channels was varied given a constant noise strength of 1.
    Error bars show the standard deviation over 3 models for handcrafted features and 6 models for neural networks.
    Using an automated noise handling method (Autoreject; second column) provided some improvement in noise robustness over using no denoising strategy at all (first column).
    Data augmentation benefited all methods, but deep learning approaches and in particular DSF (third column, in red and magenta) yielded the best performance under channel corruption.}
    \label{fig:denoising_methods_tuab}
\end{figure}

If including additional EEG channels does not by itself resolve performance degradation under channel corruption, what can be done to improve the robustness of standard EEG classification methods?
We evaluated the performance of our models when combined with three denoising strategies (Section~\ref{subsec:compared_methods}) for a fixed 6-channel montage\footnote{This 6-channel montage (Fp1, Fp2, T3, T4, Fz, Cz) performed similarly to a 21-channel montage in no-corruption conditions (Fig.~\ref{fig:impact_noise_on_n_channels}) while being more representative of the sparse montages likely to be found in mobile EEG devices.}.
Results on pathology detection (TUAB) are presented in Fig.~\ref{fig:denoising_methods_tuab}.

Without any dedicated denoising strategy, all methods showed a similarly steep performance decrease as noise became stronger (A) or more channels were corrupted (B).
Automated noise handling with Autoreject (second column) reduced differences between methods when noise strength was increased (A), and helped marginally improve robustness when only one or two channels were corrupted (B).
However, it is only with data augmentation that clear performance improvements could be obtained, allowing all methods to perform considerably better in the noisiest settings (third column).
Performance of traditional baselines was degraded however in low noise conditions.
Neural networks, in contrast, saw their performance increase the most across noise strengths and numbers of corrupted channels.
Whereas they suffered a decrease in performance of at least 34.6\% with the other strategies when going from no noise to strongest noise, training neural networks with data augmentation reduced performance loss to 5.3-10.5\% on average.
The DSF models improved performance further still over the vanilla ShallowNet by yielding an improvement of \eg 1.8-7.5\% across noise strengths.
Finally, adding the matrix logarithm and the soft-thresholding nonlinearity (DSFm-st, in magenta) yielded marginal improvements over DSFd.
Notably, under strong noise corruption ($\eta=1$) our best performing model (DSFm-st + data augmentation) yielded an accuracy improvement of 29.4\% over the vanilla net without denoising.
Overall, this suggests that learning end-to-end to both predict and handle channel corruption at the same time is key to successfully improving robustness.

\begin{figure}
    \centering
    \includegraphics[width=\textwidth]{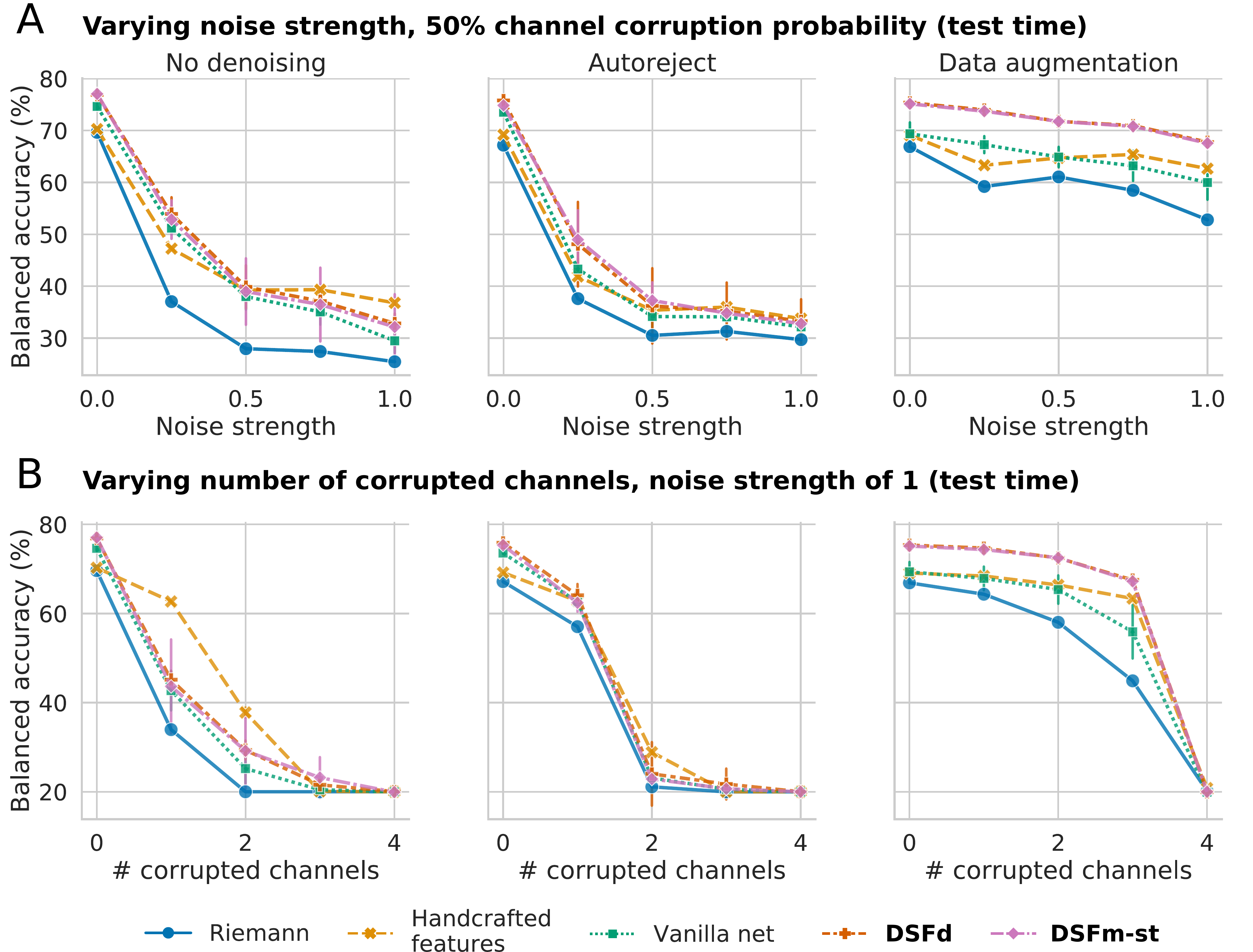}
    \caption{Impact of channel corruption on sleep staging performance for models coupled with (1) no denoising strategy, (2) Autoreject and (3) data augmentation.
    We compared the test balanced accuracy on PC18 (4-channel montage) as (A) the $\eta$ noise strength parameter was varied given a constant channel corruption probability of 50\%, and (B) the number of corrupted channels was varied given a constant noise strength of 1.
    Error bars show the standard deviation over 3 models for handcrafted features and 4 models for neural networks.
    Similarly to Fig.~\ref{fig:denoising_methods_tuab}, automated noise handling provided a marginal improvement in noise robustness in some cases, data augmentation yielded a performance boost for all methods, while a combination of data augmentation and DSF (third column, red and magenta lines which overlap) led to the best performance under channel corruption.
    }
    \label{fig:denoising_methods_pc18}
\end{figure}

Next, we repeated this analysis on a sleep staging task using the PC18 dataset (Fig.~\ref{fig:denoising_methods_pc18}).
Similarly to previous results, not using a denoising strategy led to a steep decrease in performance.
Once more, Autoreject leveled out differences between the different methods and boosted performance under single-channel corruption, but otherwise did not generally improve or degrade performance as compared to training models without a denoising strategy.
Data augmentation, in contrast, again helped improve the robustness of all methods.
Interestingly, it benefited non-deep learning approaches more clearly than was observed in pathology detection, enabling for instance the handcrafted features to reach a similar performance to the vanilla StagerNet.
DSF remained the most robust though with both DSFd and DSFm-st consistently outperforming all other methods.
Notably, the performance of these two methods was highly similar, producing mostly overlapping lines in Fig.~\ref{fig:denoising_methods_pc18}.

\begin{figure}
    \centering
    \includegraphics[width=\textwidth]{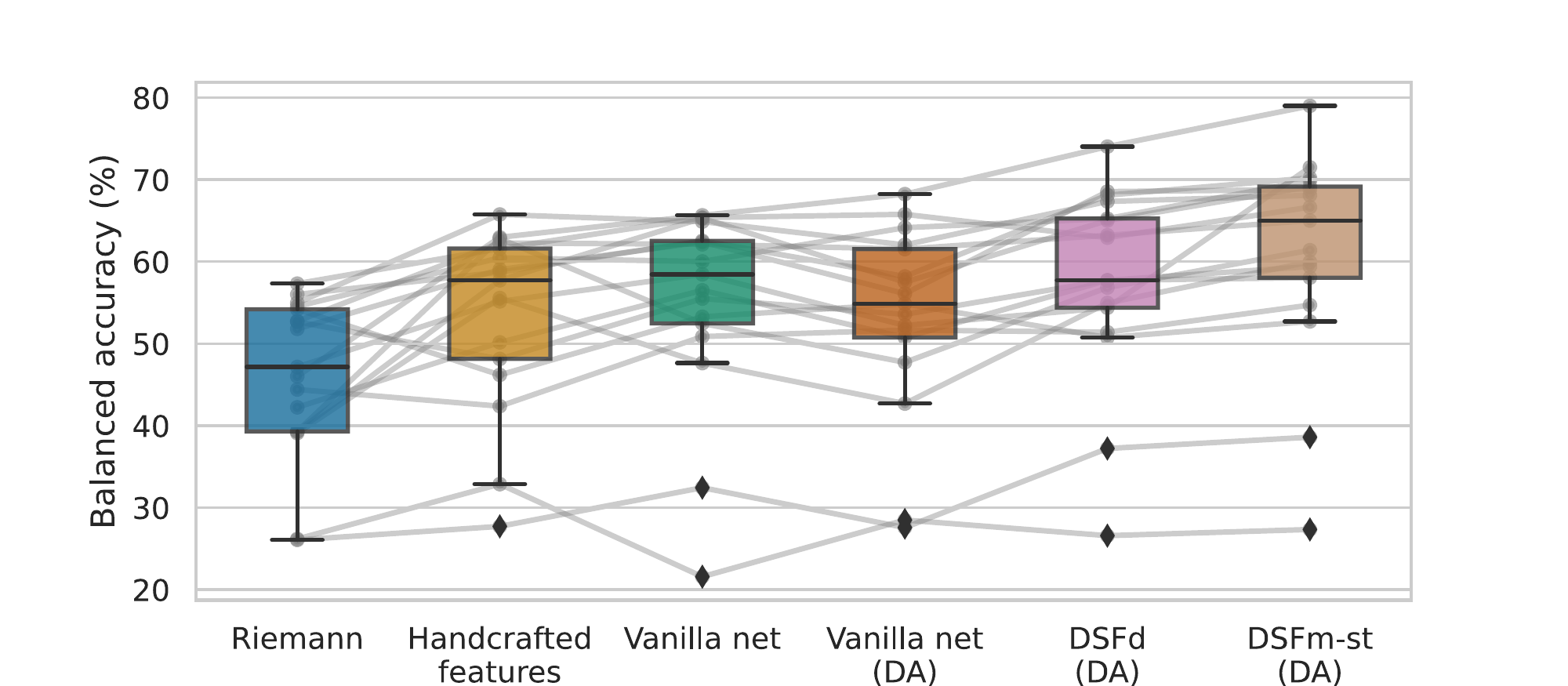}
    \caption{Recording-wise sleep staging results on MSD. Test balanced accuracy is presented for the Riemann, handcrafted features and vanilla net models without a denoising strategy, and for the vanilla net, DSFd and DSFm-st models with data augmentation (DA). Each point represents the average performance obtained by models with different random initializations (1, 3 and 9 initializations for Riemann, handcrafted features and deep learning models, respectively) on each recording from the test set of MSD. Lines represent individual recordings. The best performance was obtained by combining data augmentation with DSF with logm(cov) and soft-thresholding (DSFm-st).}
    \label{fig:denoising_methods_msd}
\end{figure}

Finally, do these results hold under more intricate, naturally occurring corruption such as found in at-home settings?
To verify this, we trained the same sleep staging models as above on the cleanest recordings of MSD (4-channel mobile EEG), and evaluated their performance on the 17 most corrupted recordings of the dataset.
Results are presented in Fig.~\ref{fig:denoising_methods_msd}.
As above, the Riemann approach did not perform well, while the handcrafted features approach was more competitive with the vanilla StagerNet without denoising.
Adding data augmentation hurt vanilla net performance on average, but once combined with dynamic spatial filters (DSFd and DSFm-st), helped improve performance over other methods.
For instance, DSFm-st with data augmentation yielded a median balanced accuracy of 65.0\%, as compared to 58.4\% for a vanilla network without denoising.
Performance improvements were as high as 14.2\% when looking at individual sessions.
Importantly, all recordings saw an increase in performance, showing the ability of our proposed approach to improve robustness in noisy settings.

Taken together, our experiments on simulated and natural channel corruption indicate that a strategy combining an attention mechanism and data augmentation yields higher robustness than traditional baselines and existing automated noise handling methods.

\subsection{Attention weights are interpretable and correlate with signal quality}
\label{subsec:attention_viz}

\begin{figure}
    \centering
    \includegraphics[width=\textwidth]{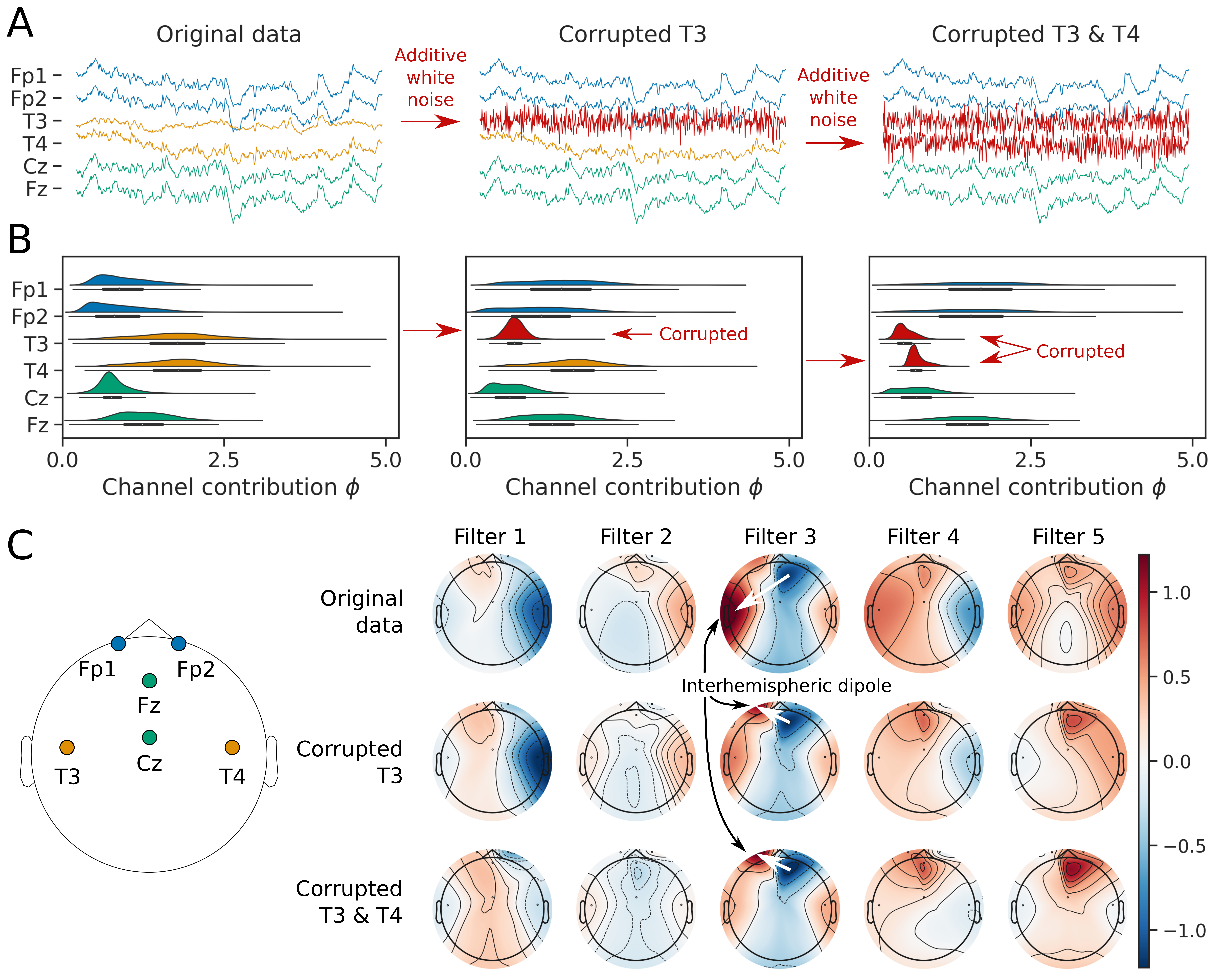}
    \caption{Channel contribution and spatial filters predicted by the DSF module trained on pathology detection.
    We compared three scenarios on the TUAB evaluation set: no added corruption, only T3 is corrupted and both T3 and T4 are corrupted.
    (A) The corruption process was carried out by replacing a channel with white noise ($\sigma \sim\mathcal{U}(20,50)$~$\mu$V), as illustrated with a single 6-s example window (first row).
    (B) The distribution of channel contribution values $\phi$ is presented using density estimate and box plots. 
    Corrupted channels are significantly down-weighted in the spatial filtering.
    (C) A subset of the spatial filters (median across all windows) are plotted as topomaps for the three scenarios.
    Corrupting T3 overall reduced the importance attributed to T3 and slightly boosted T4 values, while corrupting both T3 and T4 led to a reduction of $\phi$ for both channels, but to an increase in importance attributed to other channels.
    This change was also reflected in the overall topography: dipole-like patterns (indicated by white arrows) were dynamically modified to focus on clean channels (\eg Filter 3).
    }
    \label{fig:channel_contribution_tuab}
\end{figure}

The DSF module was key to achieving high robustness to channel corruption on both pathology detection and sleep staging tasks.
Can we explain the behavior of the module by inspecting its internal functioning?
If so, in addition to improving robustness, DSF could also be used to dynamically monitor the importance of each incoming EEG channel, providing an interesting ``free'' insight into signal quality.
To test this, we analyzed the contribution $\phi_i$ of each EEG channel $i$ to the spatial filters over the TUAB evaluation set.
Results are shown in Fig.~\ref{fig:channel_contribution_tuab}.

Overall, the attention weights behaved as could be expected: the more usable (\textit{i.e.}, noise-free) a channel was, the higher its channel importance $\phi_i$ was relative to those of other channels.
For instance, without any additional corruption, the DSF module focused most of its attention on channels T3 and T4 (Fig.~\ref{fig:channel_contribution_tuab}A, first column), known to be highly relevant for pathology detection \cite{schirrmeister2017pathology,gemein2020machine}.
However, when channel T3 was replaced with white noise, the DSF module reduced its attention to T3 and instead further increased its attention on other channels (second column).
Similarly, when both T3 and T4 were corrupted the module reduced its attention on both channels and leveraged the remaining channels instead, \textit{i.e.}, mostly Fp1 and Fp2 (third column).
Interestingly, this change is reflected by the topography of the predicted filters $W_{\mathrm{DSF}}$ (Fig.~\ref{fig:channel_contribution_tuab}B): for instance, some dipolar filters computing a difference between left and right hemispheres were dynamically adapted to rely on Fp1 or Fp2 instead of T3 or T4 (\eg filters 1, 3 and 5).
Intuitively, the network has learned to ignore corrupted data and to focus its attention on the good EEG channels, and to do so in a way that preserves the meaning of each virtual channel.

\subsection{Deconstructing the DSF module}
\label{subsec:ablation_study}

\begin{figure}
    \centering
    \includegraphics[width=0.8\textwidth]{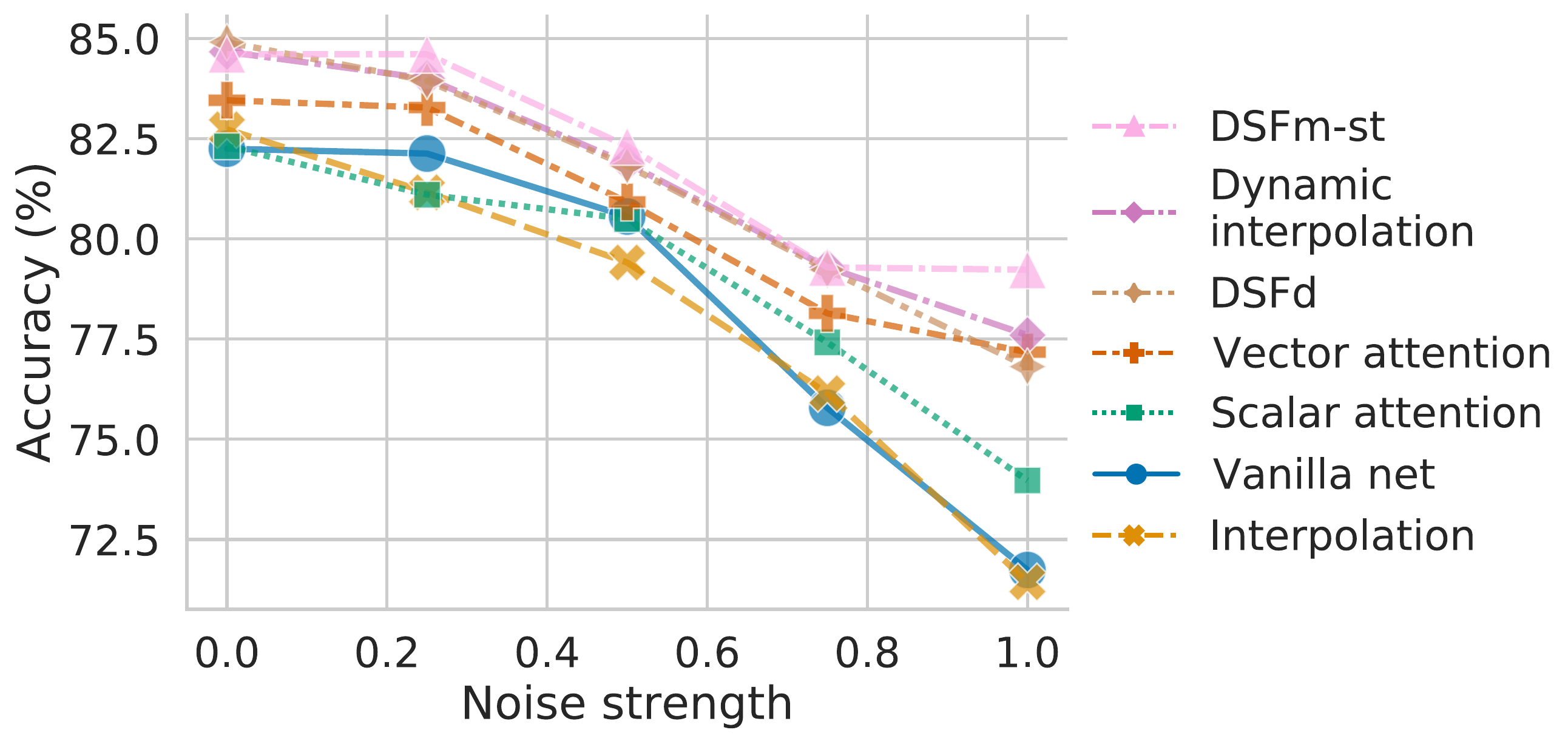}
    \caption{Performance of different attention module architectures on the TUAB evaluation set under increasing channel corruption noise strength.
    Each line represents the average of 6 models (2 random initializations, 3 random splits).
    Models that dynamically generate spatial filters, such as DSF, outperform simpler architectures across noise levels.}
    \label{fig:ablation_study_tuab}
\end{figure}

What might explain the capacity of the DSF module to improve robustness to channel corruption and provide interpretable attention weights?
By comparing DSF to simpler interpolation-based methods, DSF can be understood as a more complex version of a simple attention-based model that decides how much each input EEG channel should be replaced by its interpolated version (details provided in Appendix~\ref{sec:interp_to_dsf}).
With this interesting connection in mind, we performed an ablation study to understand the importance of each additional mechanism leading to the formulation of the DSF module.
Fig.~\ref{fig:ablation_study_tuab} shows the performance of the different attention module variations trained on the pathology detection task with data augmentation, under different noise strengths.

Naive interpolation of each channel based on the $C-1$ others (orange) performed similarly to or worse than the vanilla ShallowNet model (blue) across noise strengths.
Introducing a single attention weight (green) to control how much channels should be mixed with their interpolated version only improved performance for noise strengths above 0.5.
Using one attention weight per channel (red) further improved performance, this time across all noise strengths.
The addition of dynamic interpolation (magenta), in which both the attention weights and an interpolation matrix are generated based on the input EEG window, yielded an additional substantial performance boost.
Relaxing the constraints on the interpolation matrix and adding a bias vector to obtain DSFd (brown) led to very similar performance.
Finally, the addition of the soft-thresholding non-linearity and the use of the matrix logarithm of the covariance matrix (DSFm-st, pink) further yielded performance improvements. 

Together, these results show that combining channel-specific interpolation and dynamic prediction of interpolation matrices is necessary to outperform simpler attention module formulations.
Performance can be further improved by providing the full covariance matrix as input to the attention module and encouraging the model to produce 0-weights with a nonlinearity.  

\section{Related work}
\label{sec:related_work}

\subsection{Deep learning and noise robustness for audio data}

Noise robustness is of particular interest to the speech recognition community.
For example, ``noise-aware training'' was proposed to train deep neural networks on noisy one-channel speech signals by providing an estimate of the noise level as input to the network \cite{seltzer2013investigation}.
Combined with dropout, this approach substantially improved performance over previous state-of-the-art models.
The development of noise-invariant representations of speech signals was also investigated \cite{serdyuk2016invariant}.
This was done by training a classifier to simultaneously perform well on the speech recognition task, and badly on a domain discrimination task performed by a separate neural network head which predicted whether the input was clean or noisy.
Similarly, the ``invariant representation learning'' approach of \cite{liang2018learning,salazar2018invariant} penalized the distance between the internal representations of clean and noisy signals.
Together, these approaches improved noise robustness for single-channel speech recognition models.

Methods have also been designed to leverage the spatial information of multiple audio channels (\textit{i.e.}, an array of microphones) in a similar way to our proposed DSF approach.
Specifically, deep beamforming networks were proposed to dynamically reweight different audio channels to improve robustness to noise.
In \cite{li2016neural}, a ``filter prediction'' subnetwork was used to dynamically generate temporal filters for each of the two input channels, yielding performance and efficiency improvements over other methods.
A similar idea was developed in \cite{xiao2016deep,xiao2016beamforming}, where frequency- and channel-wise spatial filters were dynamically predicted by a subnetwork fed with spatial information such as the filter-bank spatial covariance matrices.
In a fashion similar to ours, recent work also used spatial attention to reweight statically-beamformed input speech signals to decide which filters to focus on \cite{he2020spatial}.

\subsection{Attention mechanisms for EEG processing}

Recent efforts in the deep learning and EEG community have led to various applications of attention mechanisms to end-to-end EEG processing.
Examples can be grouped in two broad categories, depending on the goal behind using attention: improving downstream performance and montage-invariant processing.

Studies leveraging attention to improve state-of-the-art performance on a specific task have used mechanisms that focus on different dimensions of an EEG representation.
For instance, attention modules taken from the natural language processing (NLP) literature were used in sleep staging architectures to improve processing of temporal dependencies \cite{phan2019seqsleepnet,yuan2019hybrid,guillot2020dreem,phan2020xsleepnet,guillot2021robust}.
In these examples, bidirectional recurrent layers were used with Luong-like attention to dynamically focus on the most important temporal slices of the input data.
Attention was also applied in the spatial dimension to allow dynamically combining information from different EEG channels \cite{yuan2018novel,yuan2019fusionatt} or even from heterogeneous channel types \cite{yuan2019hybrid}.
Likewise, the Squeeze-and-Excitation block \cite{hu2018squeeze} was used to provide both spatial and temporal attention in BCI classification tasks \cite{huang2019deep}.

Attention mechanisms have also been used to enable transfer learning between different datasets with possibly different montages. 
In \cite{nasiriattentive}, two parallel attention mechanisms allowed the neural network to focus on the channels and windows that were the most transferable when training classifiers that needed to generalize from one dataset to another.
Combined with an adversarial loss, this approach improved domain adaptation performance on a cross-dataset sleep staging task.
Similarly to our DSF approach, a spatial attention block was used in \cite{guillot2021robust} to recombine input channels into a fixed number of virtual channels and allow models to be transferred to different montages.
In the same vein, Saeed \textit{et al.} presented a Transformer-like spatial attention module to dynamically re-order input channels \cite{saeed2020learning}.
Their approach, like ours, leveraged a data augmentation transform to develop robust EEG classifiers.
In contrast to DSF, though, these approaches used attention weights in the $[0,1]$ range, breaking the conceptual connection between channel recombination and spatial filtering.

\section{Discussion}
\label{sec:discussion}
    
We introduced Dynamic Spatial Filtering (DSF), a new method to handle channel corruption in EEG based on an attention mechanism architecture and a data augmentation transform.
Plugged into a neural network whose input has a spatial dimension (\eg EEG channels), DSF predicts spatial filters that allow the model to dynamically focus on important channels and ignore corrupted ones.
DSF shares links with interpolation-based methods traditionally used in EEG processing to recover bad channels, but in contrast does not require separate preprocessing steps (often expensive when many channels are available or poorly adapted when only few channels are available).
DSF outperformed feature-based approaches and automated denoising pipelines under simulated corruption on two large public datasets and in two different predictive tasks.
Similar results were obtained on a smaller dataset of mobile sparse EEG with strong natural corruption, demonstrating the applicability of our approach to challenging at-home recording conditions.
Finally, the inner functioning of DSF can easily be inspected using a simple measure of channel importance and topographical maps.
Overall, DSF is computationally lightweight, easy to implement, and improves robustness to channel corruption in sparse EEG settings.

\subsection{Handling EEG channel loss with existing denoising strategies}

As opposed to the more general problem of ``noise handling'' which has been extensively studied in the literature (Table~\ref{tab:denoising-approaches}), we specifically focused our experiments on the problem of channel corruption in sparse montages.
In light of our results, we explain why existing strategies are not well suited for handling channel corruption, while DSF is.

Our first experiment (Section~\ref{subsec:n_channels_results}) demonstrated that adding more EEG channels does not necessarily make a classifier more robust to channel loss.
In fact, we observed the opposite: a model trained on two channels can outperform 6- and 21-channel models under heavy channel corruption (Fig.~\ref{fig:impact_noise_on_n_channels}A).
This can be explained by two phenomena.
First, increasing the number of channels increases (linearly or superlinearly\footnote{In the case of handcrafted features that look at channel pairs, \eg phase locking value.}) the input dimensionality of classifiers, making them more likely to overfit the training data.
Tuning regularization hyperparameters can help with this, but does not solve the problem by itself.
Second, in vanilla neural networks, the weights of the first spatial convolution layer, \textit{i.e.}, the spatial filters applied to the input EEG, are fixed.
This means that if one of the spatial filter outputs relies mostly on one specific important input channel, \eg T3, and this input channel is corrupted, all successive operations on the resulting virtual channel will carry noise as well.
This highlights the importance of dynamic reweighting: with DSF, we can find alternative spatial filters when an important channel is corrupted, and even completely ignore a corrupted channel if it contains no useful information.

Since adding channels is not on its own a solution, can traditional EEG denoising techniques, \eg interpolation-based methods such as Autoreject \cite{jas2017autoreject}, help handle the channel corruption problem?
In our experiments, interpolation-based denoising did help but only marginally (middle column of Fig.~\ref{fig:denoising_methods_tuab} and \ref{fig:denoising_methods_pc18}).
The relative ineffectiveness of this approach can be explained by the very low number of available channels in our experiments (4 or 6) which likely harmed the quality of the interpolation.
Our results therefore do not invalidate these kinds of methods (whose performance has been demonstrated multiple times on denser montages and in challenging noise conditions \cite{nolan2010faster,bigdely2015prep,jas2017autoreject}) but only expose their limitations when working with few channels.
Still, there are other reasons why interpolation-based methods might not be optimal in settings like the ones studied in this paper.
For instance, completely replacing a noisy channel by its interpolated version means that any remaining usable information in this channel will be discarded and that any noise contained in the other (non-discarded) channels will end up in the interpolated channel.
In addition to this, automated denoising techniques require an additional preprocessing step at both training and inference time, which adds complexity to a prediction pipeline.\footnote{This criticism applies to reconstruction-based methods as well, for which deep learning-based interpolation with \eg a GAN \cite{svantesson2020virtual} can be costly.}
An end-to-end solution such as DSF takes care of both these issues by (1) dynamically deciding how much of each channel should be used and (2) not requiring any extra steps at training or inference time.

\subsection{Impact of the input spatial representation}

The spatial representation used by the DSF model constrains the types of patterns that can be leveraged to produce spatial filters.
For instance, only using the log-variance of each channel allows detecting large-amplitude corruption or artifacts, however this makes the DSF model blind to more subtle kinds of interactions between channels.
These interactions can be very informative in certain cases, \eg when one channel is corrupted by a noise source which also affects other channels but to a lesser degree.

Our experiments suggested that models based on log-variance (DSFd) or vectorized covariance matrices (DSFm-st) were roughly equivalent in simulated noise conditions (Fig.~\ref{fig:denoising_methods_tuab}-\ref{fig:denoising_methods_pc18}).
This can be explained by the fact that the additive white noise used in these experiments was not spatially correlated and therefore there were no spatial interactions the DSF modules could have used to identify noise.
On naturally corrupted data however, using the full spatial information along with soft-thresholding was critical to outperforming other methods (Fig.~\ref{fig:denoising_methods_msd}).
This is likely because the naturally occuring noise in at-home recordings was often correlated spatially.
In addition, corrupted channels in MSD usually arose from electronic noise that completely overpowered EEG signals (Appendix~\ref{sec:msd_corruption_analysis}).
In this case, DSF could completely ignore these channels to get rid of noise.

Related attention block architectures have used average-pooling \cite{hu2018squeeze} or a combination of average- and max-pooling \cite{woo2018cbam} to summarize channels.
Intuitively, average pooling should not result in a useful representation of input EEG channels, as EEG channels are often assumed to have zero-mean, or are explicitly highpass filtered to remove their DC offset.
Max-pooling, on the other hand, does capture amplitude information that overlaps with second-order statistics, however it does not allow differentiating between large transient artifacts that only affect a small portion of a window and more temporally consistent corruption.
Experiments on TUAB (not shown) confirmed this in practice, with a combination of min- and max-pooling being less robust to noise than covariance-based models.
From this perspective, vectorized covariance matrices are an ideal choice of spatial representation for dynamic spatial filtering of EEG.
Other representations could be investigated (Appendix~\ref{sec:spatial_representation}) such as correlation matrices.
Ultimately, the DSF module could be trained on any learned representations with a spatial dimension (\eg filter-bank representations obtained with a temporal convolution layer).

\subsection{Impact of the data augmentation transform}

Data augmentation was critical to developing invariance to corruption (Section~\ref{subsec:denoising_results}).
In fact, under simulated corruption, a vanilla neural network without the DSF module gained considerable robustness once trained with our data augmentation transform, even without an attention mechanism.
Does this mean that data augmentation is the key ingredient to DSF?
In fact, our results on naturally corrupted data (Fig.~\ref{fig:denoising_methods_msd}) showed that using data augmentation without attention negatively impacted performance and that further adding an attention mechanism was necessary to improve performance.
Moreover, traditional pipelines (handcrafted features with random forests and Riemannian geometry-based models) generally did not benefit from data augmentation as much as neural networks did, and even saw their performance degrade considerably in low noise conditions in pathology detection experiments.

Nonetheless, these results highlight the role of data augmentation transforms in developing robust representations of EEG.
Recently, work in self-supervised learning for EEG \cite{banville2020uncovering,cheng2020subject,mohsenvand2020contrastive} has further suggested the importance of well-characterized data augmentation transforms for representation learning.
Ultimately, our additive white noise transform could be combined with channel masking and shuffling \cite{saeed2020learning} and other potential corruption processes such as those described in \cite{cheng2020subject,mohsenvand2020contrastive}. 

\subsection{Interpreting dynamic spatial filters to measure channel importance}
\label{subsec:channel_importance}

The results in Fig.~\ref{fig:channel_contribution_tuab} demonstrated that the spatial filters produced by the DSF module can be visualized to understand what spatial patterns a model has learned to focus on (Section~\ref{subsec:attention_viz}).
As observed in our experiments, a higher $\boldsymbol \phi$ indicates higher importance of a channel for the downstream task: for instance, temporal channels were given a higher importance in the pathology detection task, as expected from previous work on the TUAB dataset \cite{schirrmeister2017pathology,gemein2020machine}.
To further facilitate the quantitative interpretation of channel contribution, \eg in real-time settings where the user might want to monitor signal quality, a relative channel contribution metric $\boldsymbol \phi_{\mathrm{rel}} \in \left[0, 1\right]$ could be obtained by dividing $\boldsymbol \phi$ by its maximum across channels.

However, $\boldsymbol \phi$ is not a strict measure of signal quality but more of channel usefulness: there could be different reasons behind the boosting or attenuation of a channel by the DSF module.
Naturally, if a channel is particularly noisy, its contribution might be brought down to zero to avoid contaminating virtual channels with noise.
Conversely though, if the noise source behind a corrupted channel is also found (but to a lesser degree) in other channels, the corrupted channel could also be used to regress out noise and recover clean signals \cite{haufe2014interpretation}. 
In other words, $\boldsymbol \phi$ reflects the importance of a channel conditionally to others.

\subsection{Practical considerations}

When faced with possible channel corruption in a predictive task, which modelling and denoising strategies should be preferred?
This choice should depend on the number of available channels, as well as on assumptions about the stationarity of the noise.
When using sparse montages, as shown in this paper, different solutions can lead to good results.
For instance, handcrafted features with random forests can perform well when spatial information is not critical (\eg sleep staging, Section~\ref{subsec:denoising_results}) or noise is stationary \cite{engemann2018robust}, although they require a non-trivial feature engineering step.
However, when less can be assumed about the predictive task, \eg corruption might be non-stationary or spatial information is likely important, DSF with data augmentation is an effective way to make a neural network corruption-robust.
Although we did not specifically test denoising approaches on dense montages (\eg above six channels), we can expect different methods to work well in these settings. 
For instance, under stationary noise, Riemmanian geometry-based approaches were shown to be robust to the lack of preprocessing in MEG data \cite{sabbagh2020predictive}.
If, on the other hand, noise is not stationary and the computational resources allow it, interpolation-based methods might be used to impute missing channels before applying a predictive model (\eg \cite{jas2017autoreject}).
In cases where introducing a separate preprocessing step is not desirable and especially when using differentiable programming models, DSF with data augmentation might again be a promising end-to-end solution.\footnote{In this case, the number of parameters of the module can be controlled by \eg selecting log-variance as the input representation or reducing dimensionality by using fewer spatial filters than there are input channels.}

\subsection{Limitations}

Our experiments on sleep data focused on window-wise decoding, \textit{i.e.,} we did not aggregate larger temporal context with recurrent layers, but directly mapped each window to a prediction.
However, modeling these longer-scale temporal dependencies was recently shown to help sleep staging performance significantly \cite{supratak2017deepsleepnet,chambon2018deep,phan2019seqsleepnet,yuan2019hybrid,guillot2020dreem,phan2020xsleepnet,guillot2021robust}.
Despite a slight decrease in final performance of our sleep staging models, window-wise decoding offered a simple but realistic setting to test robustness to channel corruption, 
while limiting the number of hyperparameters and the computational cost of the experiments.
Interestingly, our pathology detection results (in which temporal aggregation was done by averaging predictions over an entire recording) remained only a few accuracy percentage points below the state of the art, although we used a relatively simple architecture and focused on only 6 of the 21 channels typically used. In practice, the effect of data corruption by far exceeded the drop in performance caused by using slightly simpler architectures.

The data augmentation and the noise corruption strategies exploited in this work employ additive Gaussian white noise. While this approach helped develop noise robust models,
non-spatially correlated additive white noise represents an ``adversarial scenario''. Indeed,
under strong white noise, the information in higher frequencies is more likely to be lost than with \eg pink or brown noise. Additionally, the absence of spatial noise correlation means that spatial filtering can less easily leverage multi-channel signals to regress out noise (Section~\ref{subsec:channel_importance}).
Exploring more varied and realistic types of channel corruption could further help clarify the ability of DSF to work under different conditions.
Despite this, our experiments on naturally corrupted sleep data showed that additive white noise as a data augmentation does help improve robustness under noisy conditions.

Finally, we focused our empirical study of channel corruption on two clinical problems that are prime contenders for mobile EEG applications: pathology screening and sleep monitoring. 
Interestingly, these two tasks have been shown to work well even with limited spatial information (\textit{i.e.}, single-channel sleep staging \cite{liang2012automatic}) or to be highly correlated with simpler spectral power representations \cite{schirrmeister2017pathology}.
Therefore, future work will be required to validate the use of DSF on tasks where fine-grained spatial patterns might be critical to successful prediction, \eg brain age estimation \cite{engemann2020combining}.

\section{Conclusion}

We presented Dynamic Spatial Filtering (DSF), an attention mechanism architecture that improves robustness to channel corruption in EEG prediction tasks.
Combined with a data augmentation transform, DSF outperformed other noise handling procedures under simulated and real channel corruption on three datasets.
Moreover, DSF enables efficient end-to-end handling of channel corruption, works with few channels, is interpretable and does not require expensive preprocessing.
We hope that our method can be a useful tool to improve the reliability of EEG processing in challenging non-traditional settings such as user-administered, at-home recordings.

\section*{Data and code availability statement}
The TUAB and PC18 datasets are openly available online
\footnote{\url{https://www.isip.piconepress.com/projects/tuh_eeg/html/downloads.shtml}}
\footnote{\url{https://physionet.org/content/challenge-2018/1.0.0/}}.
MSD recordings were collected on users of the Muse S headband according to InteraXon's privacy policy\footnote{\url{https://choosemuse.com/legal/privacy/}}.
According to this policy, recordings cannot be shared unless a formal data sharing agreement has been put in place with InteraXon Inc.
Code used for data analysis can be found at \\ \url{https://github.com/hubertjb/dynamic-spatial-filtering}.

\section*{Acknowledgements}
This work was supported by Mitacs (project number IT14765) and InteraXon Inc. (graduate funding support) for HB and by the BrAIN (ANR-19-DATA-0023) and AI-cog (ANR-20-IADJ-0002) ANR grants for AG and DAE.

This work was performed using HPC resources from GENCI-IDRIS.

We would like to thank the Python \cite{rossum1995python,oliphant2007python} community for developing many of the tools used in the making of this article: NumPy \cite{numpy20}, SciPy \cite{virtanen2020scipy}, matplotlib \cite{hunter2007matplotlib}, seaborn \cite{waskom2020seaborn}, pandas \cite{mckinney2010data}, scikit-learn \cite{pedregosa2011scikit}, MNE-Python \cite{gramfort2014mne}, PyTorch \cite{paszke2019pytorch}, hydra \cite{Yadan2019Hydra}, pyRiemann \cite{BARACHANT2013172} and braindecode \cite{schirrmeister2017deep}.

\section*{Declarations of interest}
HB receives graduate funding support from InteraXon Inc.
SUNW and CA are employees of InteraXon Inc.

\begin{appendices}

\section{Representation of spatial information in the DSF module}
\label{sec:spatial_representation}

In this section, we discuss different spatial representations of EEG that can be used as input to a spatial attention block such as the DSF module.
Specifically, we consider the spatial covariance matrix along with different vectorization schemes.

Given some EEG signals $X \in \bbR^{C \times T}$, where $T$ is the number of time samples in $X$, and which we assume to be zero-mean, an unbiased estimate of their covariance reads:
\begin{equation}
    \Sigma(X) = \frac{X X^\top}{T - 1} \in \bbR^{C \times C} \enspace .
\end{equation}

The zero-mean assumption is justified after some high-pass filtering or simple baseline correction of the signals. To assess whether one channel is noisy or not, a human expert annotator will typically rely on the power of a signal and its similarity with the neighboring channels.
This information is encoded in the covariance matrix.

Multiple well-established signal processing techniques rely on some estimate of $\Sigma$.
For instance, common spatial patterns (CSP) performs generalized eigenvalue decomposition of covariance matrices to identify optimal spatial filters for maximizing the difference between two classes \cite{koles1990spatial}.
Riemannian geometry approaches to EEG classification and regression instead leverage the geometry of the space of symmetric positive definite (SPD) matrices to develop geometry-aware metrics. 
They are used to average and compare covariance matrices, which has been shown to outperform other classical approaches \cite{congedo2017riemannian,sabbagh2020predictive}.
Artefact handling pipelines such as the Riemannian potato \cite{barachant2013riemannian} and Artifact Subspace Reconstruction \cite{mullen2015real} further rely on covariance matrices to identify bad epochs or attenuate noise.

The values in a covariance matrix often follow a heavy-tailed distribution.
Therefore, knowing that neural networks are typically easier to train when the distribution of input values is fairly concentrated, it is helpful to standardize the covariance values before feeding them to the network.
While scalar non-linear transformations (\eg logarithms) could help reduce the range of values and facilitate a neural network's task, the geometry of SPD matrices actually calls for metrics that respect the Riemannian structure of the SPD matrices' manifold \cite{lin2019riemannian}.
For instance, this means using the matrix logarithm instead of naively flattening the upper triangle and diagonal of the matrix~\cite{sabbagh2020predictive}.
For an SPD matrix $S$, whose orthogonal eigendecomposition reads $S = U \Lambda U^\top$, where $\Lambda = \operatorname{diag}(\lambda_1, ..., \lambda_n)$ contains its eigenvalues, the matrix logarithm $\log(S)$ is given by:
\begin{equation}
    \label{eq:logm}
    \log(S) = U \operatorname{diag}(\log (\lambda_1), ..., \log (\lambda_n)) U^\top  \enspace .
\end{equation}

The diagonal and upper-triangular part of $\log(S)$ can then be flattened into a vector with $C(C+1)/2$ values, which is then typically used with linear models, \eg support vector machines (SVM) or logistic regression.

Other options to provide input values in a restricted range exist.
For instance, one could simply use the element-wise logarithm of the diagonal of the covariance matrix, \ie the log-variance of the input signals. This is appropriate if pairwise inter-channel covariance information is deemed not critical down the line.
Alternatively, Pearson's correlation matrix, which can be seen as the covariance matrix of the z-score normalized signals, could be used. It has the advantage that its values are already in a well-defined range (-1, 1), yet it is blind to channel variances.
In our experiments, we focused on two spatial representations: the channel-wise variance obtained from the diagonal  of $\Sigma$, and the matrix logarithm of $\Sigma$. Both helped improve robustness on the pathology detection and sleep staging tasks.

\section{Deep learning architectures}
\label{sec:architectures}

\begin{figure}
    \centering
    \includegraphics[width=\textwidth]{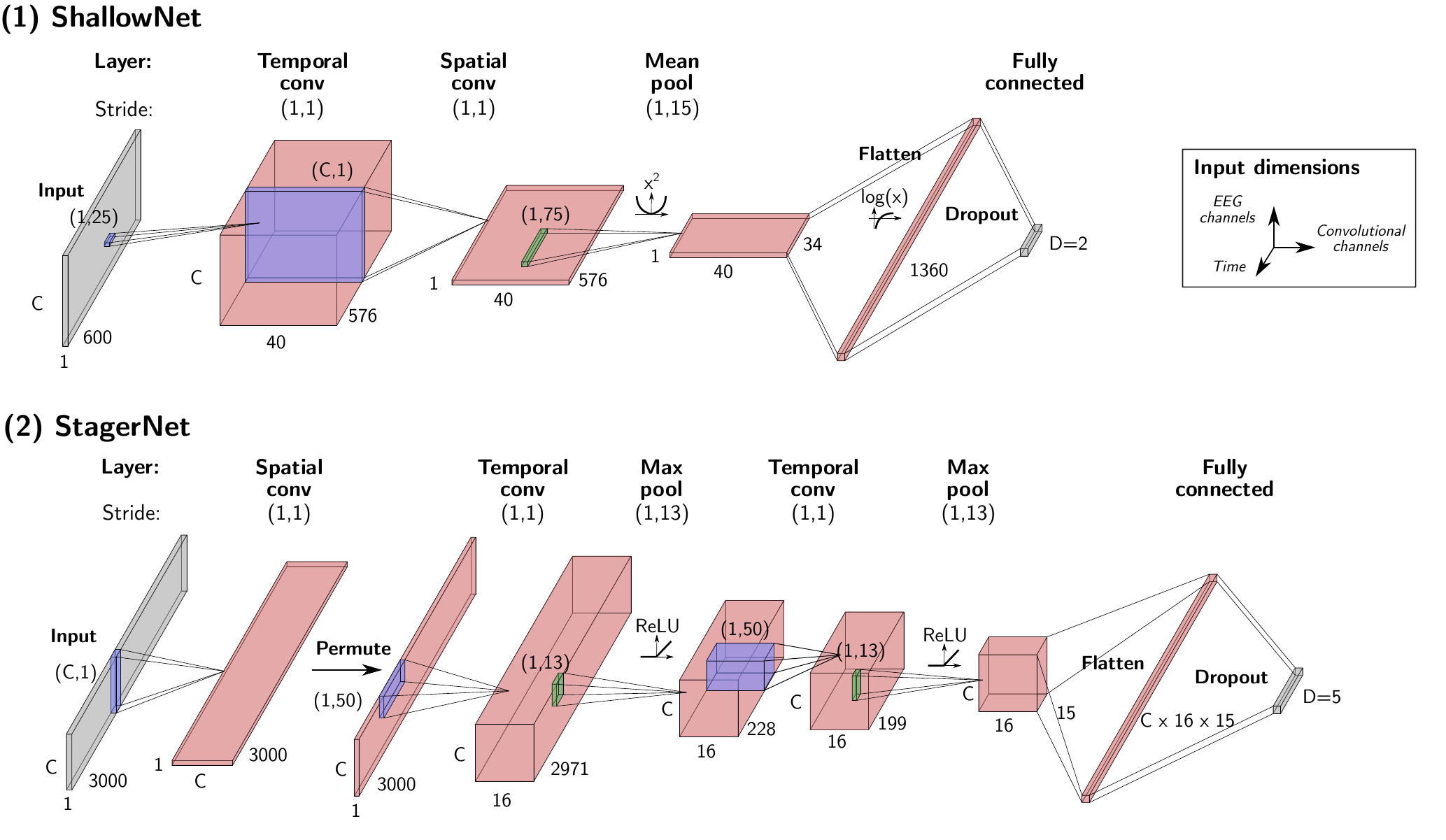}
    \caption{Neural network architectures $f_{\Theta}$ used in (1) pathology detection and (2) sleep staging experiments.}
    \label{fig:architectures}
\end{figure}

The ConvNets $f_\Theta$ used in our experiments are described in more detail in Fig.~\ref{fig:architectures}.
In MSD experiments, the input sampling rate was of 128~Hz instead of 100~Hz as for PC18. 
Therefore, we adapted the temporal convolution and max pooling hyperparameters so that they would cover approximately the same duration: filter size of 64 samples, padding size of 13 and max pooling size of 16 (vs. 50, 10 and 13, respectively).
This yielded a total of 21,369 parameters.

\section{Hyperparameter optimization of baseline models}
\label{sec:hp_opt}

A grid-search over hyperparameters of the random forest (RF) and logistic regression classifiers was performed with 3-fold cross-validation on combined training and validation sets.
This search was performed for each reported experimental configuration: for each number of channels (for experiments in Section~\ref{subsec:n_channels_results}), each denoising strategy (no denoising, Autoreject and data augmentation) and each dataset (TUAB, PC18 and MSD).

For all RF models, we used 300 trees. This turned out to be a good trade-off between model performance and computational costs.
For each experiment, we selected by cross-validation the depth of the trees among \{13,15,17,19,21,23,25\}, the split criterion between Gini and entropy, and the fraction of selected features used in each tree among `sqrt` (the square-root of the number of features is used) , `log2' (the logarithm in base 2 of the number of features is used), and using all features.
For logistic regression models, the regularization parameter $C$ was chosen among $\{10^{-4},10^{-3},\dots,10\}$.
We expanded the search on MSD as performance did not peak in the ranges considered above by adding the following values to the search space: depth in \{1,3,5,7,9,11\} and $C$ in $\{10^2,10^3,10^4,10^5\}$.

The selected hyperparameter configurations are listed in Tables~\ref{tab:hp_search_exp1} and \ref{tab:hp_search_exp2} for the experiments in Sections~\ref{subsec:n_channels_results} and \ref{subsec:denoising_results}, respectively.
Once the best hyperparameters for an experimental configuration were identified, the training and validation sets were combined into a single set on which the model with the best hyperparameters was finally trained.

\begin{table}[]
\centering
\caption{Selected hyperparameters for experiments on number of channels (Section~\ref{subsec:n_channels_results}).}
\label{tab:hp_search_exp1}
\begin{tabular}{@{}lllll@{}}
\toprule
 &  & \multicolumn{3}{c}{\textbf{Number of channels}} \\
\textbf{Model} & \multicolumn{1}{c}{\textbf{Hyperparameter}} & \multicolumn{1}{c}{\textbf{2}} & \multicolumn{1}{c}{\textbf{6}} & \multicolumn{1}{c}{\textbf{21}} \\
\midrule
Random Forest (RF) & Number of trees & 300 & 300 & 300 \\
 & Tree depth & 17 & 21 & 19 \\
 & Criterion & entropy & Gini & entropy \\
 & Features & all & all & all \\
Logistic regression (LR) & C & 0.1 & 0.1 & 0.001 \\
\bottomrule
\end{tabular}
\end{table}

\begin{table}[]
\caption{Selected hyperparameters for experiments on denoising strategies (Section~\ref{subsec:denoising_results}).}
\label{tab:hp_search_exp2}
\begin{tabular}{@{}llllll@{}}
\toprule
 &  &  & \multicolumn{3}{c}{\textbf{Denoising strategy}} \\
\textbf{Dataset} & \multicolumn{1}{c}{\textbf{Model}} & \multicolumn{1}{c}{\textbf{Hyperparameter}} & \multicolumn{1}{c}{\textbf{No denoising}} & \multicolumn{1}{c}{\textbf{Autoreject}} & \multicolumn{1}{c}{\textbf{Data augmentation}} \\ \midrule
\multirow{5}{*}{TUAB} & \multirow{4}{*}{RF} & Number of trees & 300 & 300 & 300 \\
 &  & Tree depth & 21 & 13 & 17 \\
 &  & Criterion & Gini & entropy & entropy \\
 &  & Features & all & all & all \\
 & LR & C & 0.1 & 0.1 & 0.01 \\ \midrule
\multirow{5}{*}{PC18} & \multirow{4}{*}{RF} & Number of trees & 300 & 300 & 300 \\
 &  & Tree depth & 15 & 15 & 17 \\
 &  & Criterion & entropy & Gini & entropy \\
 &  & Features & sqrt & sqrt & sqrt \\
 & LR & C & 1 & 1 & 10 \\ \midrule
\multirow{5}{*}{MSD} & \multirow{4}{*}{RF} & Number of trees & 300 & 300 & 300 \\
 &  & Tree depth & 9 & 9 & 11 \\
 &  & Criterion & entropy & entropy & entropy \\
 &  & Features & all & sqrt & sqrt \\
 & LR & C & 0.1 & 0.1 & $10^5$ \\ \bottomrule
\end{tabular}
\end{table}

\section{Analysis of channel corruption in the Muse Sleep Dataset}
\label{sec:msd_corruption_analysis}

The Muse Sleep Dataset (MSD) is a collection of at-home overnight recordings.
Therefore, sources of noise are more common in this dataset than in typical sleep datasets collected under controlled laboratory conditions (\eg PC18).

\begin{figure}
    \centering
    \includegraphics[width=0.5\textwidth]{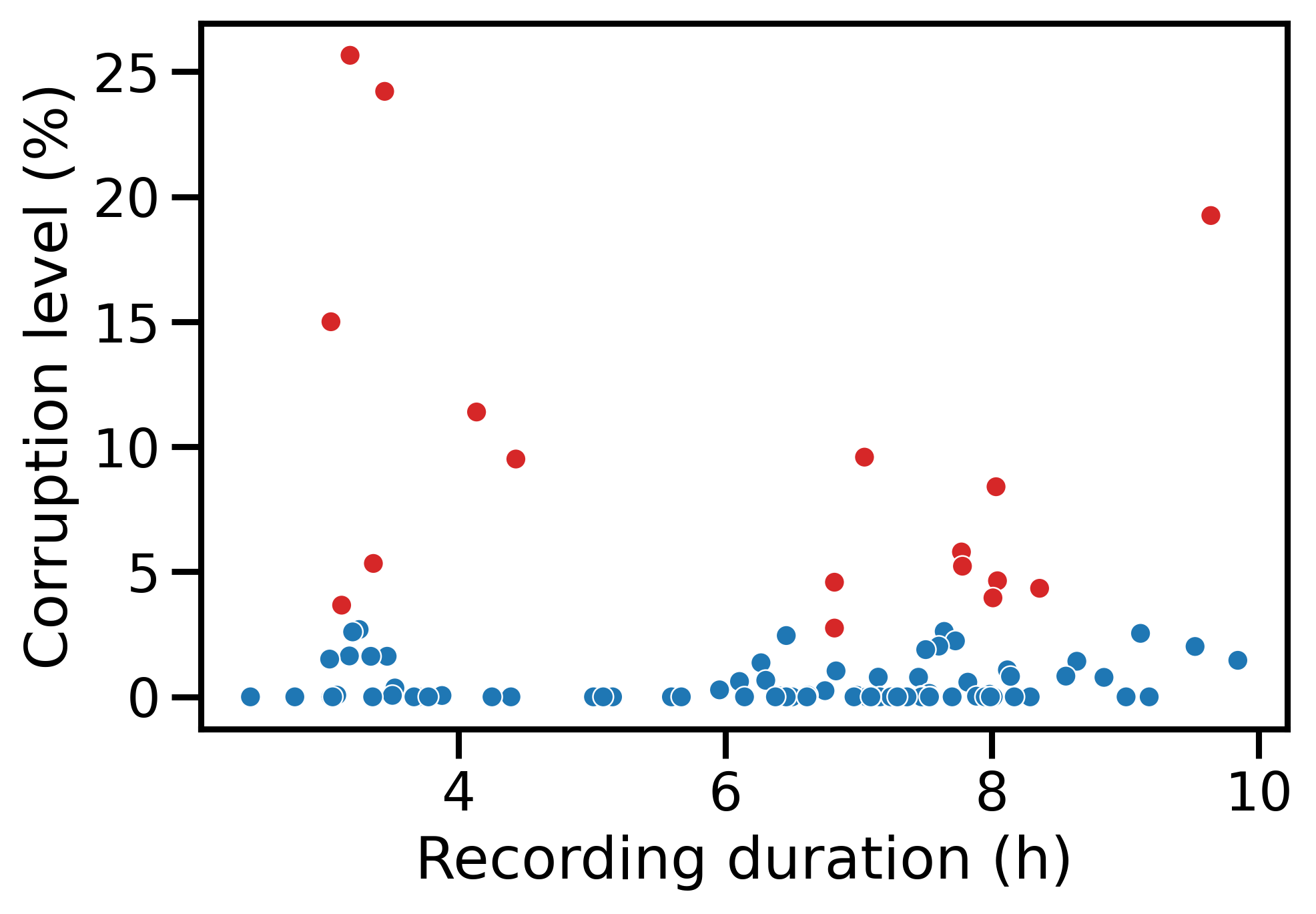}
    \caption{Corruption percentage of the 98 recordings of MSD. Each point represents a single recording. The 17 most corrupted recordings (red) were used as test set in our experiments of Section~\ref{subsec:denoising_results}.}
    \label{fig:msd_corruption_level}
\end{figure}

To characterize the prevalence of channel corruption in MSD recordings, we inspected the variance and the slope of the power spectral density (PSD) of each EEG channel across 30-s windows.
Variance is a good measure of signal quality (for instance, the DSFd variant received log-variance as input in our experiments), while the spectral slope is a global descriptor of the frequency content of a signal and allows distinguishing between channel corruption (which yields flatter spectra) and artifacts (often displaying strong low frequencies, \eg eye movements). 
Simple thresholds set empirically on these two markers allowed approximate detection of channel corruption events.
Specifically, we flagged a channel in a window as ``corrupted'' if its $\log_{10}$-$\log_{10}$ spectral slope \cite{schiratti2018ensemble} between 0.1 and 30~Hz was above -0.5 (unitless) and its variance was above 1,000~$\mu V^2$.
We then computed a single channel corruption metric per recording by computing the fraction of windowed channels which were flagged as corrupted.
About two-thirds of the recordings had no channel corruption according to this metric, while the remaining had a value of up to 25\% (Fig.~\ref{fig:msd_corruption_level}).

For our experiments on MSD, we therefore selected the 81 cleanest recordings (\textit{i.e.}, with the lowest corruption fraction) for training and validation and kept the 17 noisiest recordings for testing.
This procedure allowed testing whether a model trained on relatively clean data could perform well even when random channel corruption was introduced at inference time.

\section{From simple interpolation to Dynamic Spatial Filtering}
\label{sec:interp_to_dsf}

In this section, we establish a conceptual link between DSF and noise handling pipelines such as Autoreject (Section~\ref{subsec:sota_denoising}) which rely on an interpolation step to reconstruct channels that have been identified as bad.
Specifically, these pipelines use head geometry-informed interpolation methods (based on the 3D coordinates of EEG electrodes and spline interpolation) to compute the weights necessary to interpolate each channel using a linear combination of the $C-1$ other channels \cite{perrin1989spherical}.
From this perspective, a naive method of handling corrupted channels might be to always replace each input EEG channel by its interpolated version based on the other $C-1$ channels.
An ``interpolation-only'' module $m_{\mathrm{interp}}$ could be written as:

\begin{equation}
    m_{\mathrm{interp}}(X) = W_{\mathrm{interp}} X \enspace ,
\end{equation}
where $W_{\mathrm{interp}}$ is a $C \times C$ real-valued matrix with a 0-diagonal\footnote{$W_{\mathrm{interp}}$ can be set or initialized using head geometry information \cite{perrin1989spherical} or can be learned from the data end-to-end.}.
The limitation of this approach is that given at least one corrupted channel in the input $X$, the interpolated version of all non-corrupted channels will be reconstructed in part from corrupted channels.
This means noise will still be present, however given enough clean channels, its impact might be mitigated.

Improving upon the naive interpolation-only approach, we might add the ability for the model to decide whether (and to what extent) channels should be replaced by their interpolated version.
For instance, if the channels in a given window are mostly clean, it might be desirable to keep the initial channels; however, if the window is overall corrupted, it might instead be better to replace channels with their interpolated version.
This leads to a ``scalar-attention'' module $m_{\mathrm{scalar}}$:
\begin{equation}
    m_{\mathrm{scalar}}(X) = \alpha_X X + (1-\alpha_X) W X \enspace ,
\end{equation}
where $\alpha_X \in [0,1]$ is the attention weight predicted by an MLP conditioned on $X$ (\eg on its covariance matrix) and $W$ is the same as for the interpolation-only module.
While this approach is more flexible, it still suffers from the same limitation as before: there is a chance interpolated channels will be reconstructed from noisy channels.
Moreover, the fact that the attention weight is applied globally, \textit{i.e.}, a single weight applies to all $C$ channels, limits the ability of the module to focus on reconstructing corrupted channels only.

Instead, the ``vector attention'' module $m_{\mathrm{vector}}$ introduces channel-wise attention weights, so that the interpolation can be independently controlled for each channel:
\begin{equation}
    m_{\mathrm{vector}}(X) = \operatorname{diag}(\boldsymbol{\alpha}_X) X + (I - \operatorname{diag}(\boldsymbol{\alpha}_X)) W X \enspace ,
\end{equation}
where $\boldsymbol{\alpha}_X \in [0, 1] ^ C$ is again obtained with an MLP and $W$ is as above.
Although more flexible, this version of the attention module still faces the same problem caused by static interpolation weights.

To solve this issue, we build on the previous approach by both predicting an attention vector $\boldsymbol{\alpha}_X$ as before and dynamically interpolating with a matrix $W_X \in \bbR ^ {C \times C}$ (with a 0-diagonal) predicted by another MLP:

\begin{equation}
\label{eq:dyn_interp}
    m_{\mathrm{dynamic}}(X) = \operatorname{diag}(\boldsymbol{\alpha}_X)X + (I - \operatorname{diag}(\boldsymbol{\alpha}_X)) W_X X \enspace .
\end{equation}

In practice, a single MLP can output $C \times C$ real values, which are then reorganized into a 0-diagonal interpolation matrix $W$ and a $C$-length vector whose values are passed through a sigmoid nonlinearity to obtain the attention weights $\boldsymbol{\alpha}_X$.
An interesting property of this formulation which holds for $m_{\mathrm{vector}}$ too is that $\boldsymbol{\alpha}_X$ can be directly interpreted as the level to which each channel is replaced by its interpolated version.
However, in contrast to $m_{\mathrm{vector}}$ the interpolation filters can dynamically adapt to focus on the most informative channels.

Finally, we observe that Eq.~\eqref{eq:dyn_interp} can be rewritten as a single matrix product:
\begin{equation}
    m_{\mathrm{dynamic}}(X) = \left(\operatorname{diag}(\boldsymbol{\alpha}_X) + (I - \operatorname{diag}(\boldsymbol{\alpha}_X)) W_X \right) X = \Omega_X X \enspace ,
\end{equation}
where, denoting the element $i,j$ of matrix $W_X$ as $W_{ij}$,
\begin{equation}
    \Omega_X = \begin{bmatrix}
    \alpha_1 & (1 - \alpha_1) W_{12} & \hdots & (1 - \alpha_1) W_{1C}\\
    (1 - \alpha_2) W_{21} & \alpha_2 & \hdots & (1 - \alpha_2) W_{2C}\\
    \vdots & \vdots & \ddots & \vdots\\
    (1 - \alpha_C) W_{C1} & (1 - \alpha_C) W_{C2} & \hdots & \alpha_C\\
    \end{bmatrix} \enspace .
\end{equation}
The matrix $\Omega_X$ contains $C^2$ free variables, that are all conditioned on X through an MLP.
We can then relax the constraints on $\Omega_X$ to obtain a simple matrix $W_{\mathrm{DSF}}$ where there are no dependencies between the parameters of a row and the diagonal elements are allowed to be real-valued.
This new unconstrained formulation can be interpreted as a set of spatial filters that perform linear combinations of the input EEG channels.
We can further introduce an additional bias term to recover the DSF formulation introduced in Section~\ref{subsec:dsf_description}:
\begin{equation}
    m_{\mathrm{DSF}}(X) = W_{\mathrm{DSF}}(X) X + b_{\mathrm{DSF}}(X) \enspace .
\end{equation}
This bias term can be interpreted as a dynamic re-referencing of the virtual channels.
In contrast to the interpolation-based formulations, DSF allows controlling the number of ``virtual channels'' $C'$ to be used in the downstream neural network in a straightforward manner (\eg enabling the use of montage-specific DSF heads that could all be plugged into the same $f_\Theta$ with fixed input shape).
As shown in Section~\ref{subsec:ablation_study}, DSF also outperformed interpolation-based formulations in our experiments.

\end{appendices}

\bibliographystyle{unsrt}
\bibliography{bibli}  

\end{document}